\documentclass[conference]{IEEEtran}
\IEEEoverridecommandlockouts
\usepackage{cite}
\usepackage{amsmath,amssymb,amsfonts}
\usepackage{textcomp}
\def\BibTeX{{\rm B\kern-.05em{\sc i\kern-.025em b}\kern-.08em
    T\kern-.1667em\lower.7ex\hbox{E}\kern-.125emX}}

\usepackage[skip=4pt,font=small,labelformat=simple]{subcaption}
\usepackage{adjustbox}
\usepackage{graphicx}
\usepackage{multirow}
\usepackage{mathrsfs}
\usepackage{amsmath}
\usepackage{mathtools}
\usepackage{amsthm}
\usepackage{algpseudocodex}
\usepackage{flushend}
\usepackage{hyperref}
\usepackage{caption}
\usepackage{array}
\usepackage{cleveref}
\usepackage{algorithm}
\usepackage{enumitem}
\theoremstyle{plain}
\newtheorem{theorem}{Theorem}[section]

\newtheorem{lemma}[theorem]{Lemma}
\newtheorem{corollary}[theorem]{Corollary}
\theoremstyle{definition}
\newtheorem{definition}[theorem]{Definition}

\theoremstyle{remark}

\usepackage{stfloats}
\usepackage[numbers]{natbib}
\usepackage[textsize=tiny]{todonotes}

\newcommand{\rkat}[1]{\colorbox{gray!60}{#1}}

\newcommand{\rkbt}[1]{\colorbox{gray!30}{#1}}

\newcommand{\rkct}[1]{\colorbox{gray!15}{#1}}
\newcommand{\ds}[1]{\textls[-60]{\textsc{\MakeLowercase{#1}}}}

\newcolumntype{P}[1]{>{\centering\arraybackslash}p{#1}}
\newcolumntype{L}[1]{>{\raggedright\arraybackslash}p{#1}}

\usepackage[utf8]{inputenc} % allow utf-8 input
\usepackage[T1]{fontenc}    % use 8-bit T1 fonts
\usepackage{hyperref}       % hyperlinks
\usepackage{url}            % simple URL typesetting
\usepackage{booktabs}       % professional-quality tables
\usepackage{amsfonts}       % blackboard math symbols
\usepackage{nicefrac}       % compact symbols for 1/2, etc.
\usepackage{microtype}      % microtypography
\usepackage{xcolor}         % colors
\usepackage{xspace}
\newcommand{\mypara}[1]{
    \vspace{3pt}
    \par\noindent\textbf{#1}
    \noindent}
\NewDocumentCommand{\leo}{mg}{\IfNoValueTF{#2}
{\textcolor{teal}{#1}}
{\textcolor{teal}{$\blacktriangleright$}#1 \textcolor{teal}{$\triangleright$ #2$\blacktriangleleft$}}}
\newcommand{\aggname}{SIGMA}
\usepackage{tcolorbox}

\tcbset{colframe=white!75!black, boxrule=0.3mm, left=0.4mm, right=0.4mm, top=0.4mm, bottom=0.4mm, arc=0.75mm, after skip=1mm, before skip=2mm}
\renewcommand{\thetable}{\Roman{table}}

\begin{document}

\title{SIGMA: An Efficient Heterophilous Graph Neural
Network with Fast Global Aggregation}

\author{
\IEEEauthorblockN{Haoyu Liu, Ningyi Liao, Siqiang Luo\IEEEauthorrefmark{1}}
\IEEEauthorblockA{
\textit{College of Computing and Data Science, Nanyang Technological University}\\
haoyu.liu@ntu.edu.sg,
liao0090@e.ntu.edu.sg,
siqiang.luo@ntu.edu.sg}
\thanks{\IEEEauthorrefmark{1}Siqiang Luo is the corresponding author.}
}

\maketitle

\begin{abstract}
Graph neural networks (GNNs) realize great success in graph learning but suffer from performance loss when meeting heterophily, i.e. neighboring nodes are dissimilar, due to their local and uniform aggregation. 
Existing attempts of heterophilous GNNs incorporate long-range or global aggregations to distinguish nodes in the graph. However, these aggregations usually require iteratively maintaining and updating full-graph information, which limits their efficiency when applying to large-scale graphs. 
In this paper, we propose \aggname{}, an efficient global heterophilous GNN aggregation integrating the structural similarity measurement SimRank. Our theoretical analysis illustrates that \aggname{} inherently captures distant global similarity even under heterophily, that conventional approaches can only achieve after iterative aggregations. Furthermore, it enjoys efficient one-time computation with a complexity only linear to the node set size $\mathcal{O}(n)$.
Comprehensive evaluation demonstrates that \aggname{} achieves state-of-the-art performance with superior aggregation and overall efficiency. Notably, it obtains 5$\times$ acceleration on the large-scale heterophily dataset \emph{pokec} with over 30 million edges compared to the best baseline aggregation. 
\end{abstract}


\section{Introduction}
\label{sec:introduction}

\begin{figure*}[!t]
    \centering
%\vspace{-0.5ex}
    \subcaptionbox{Intuitive Example of Global Similarity\label{ffig:simrank_diag}}%
    [0.38\linewidth]{\includegraphics[height=1.33in]{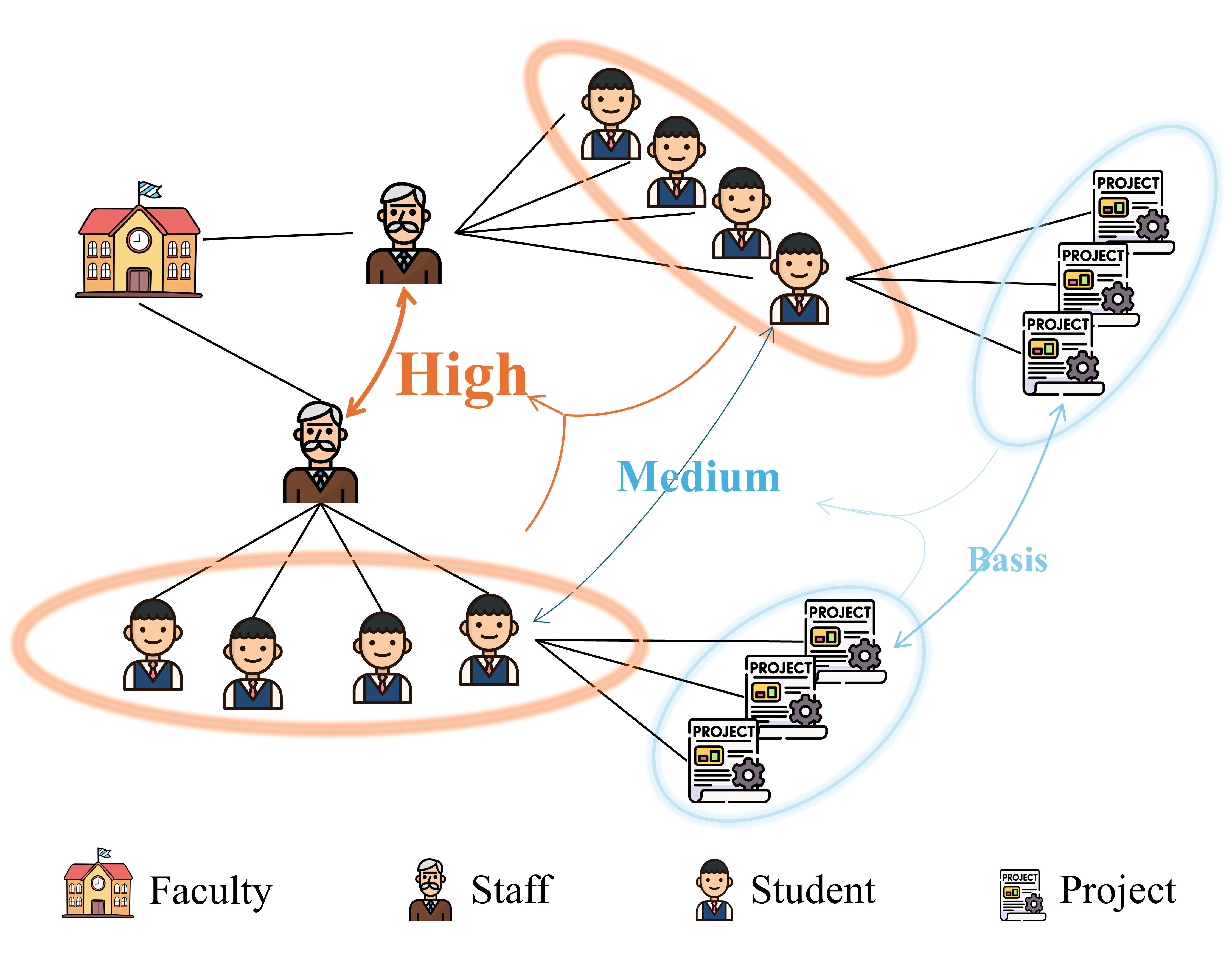}}
    \hfil
    \subcaptionbox{Local AGG\label{ffig:pagerank}}%
    [0.30\linewidth]{\includegraphics[height=1.33in]{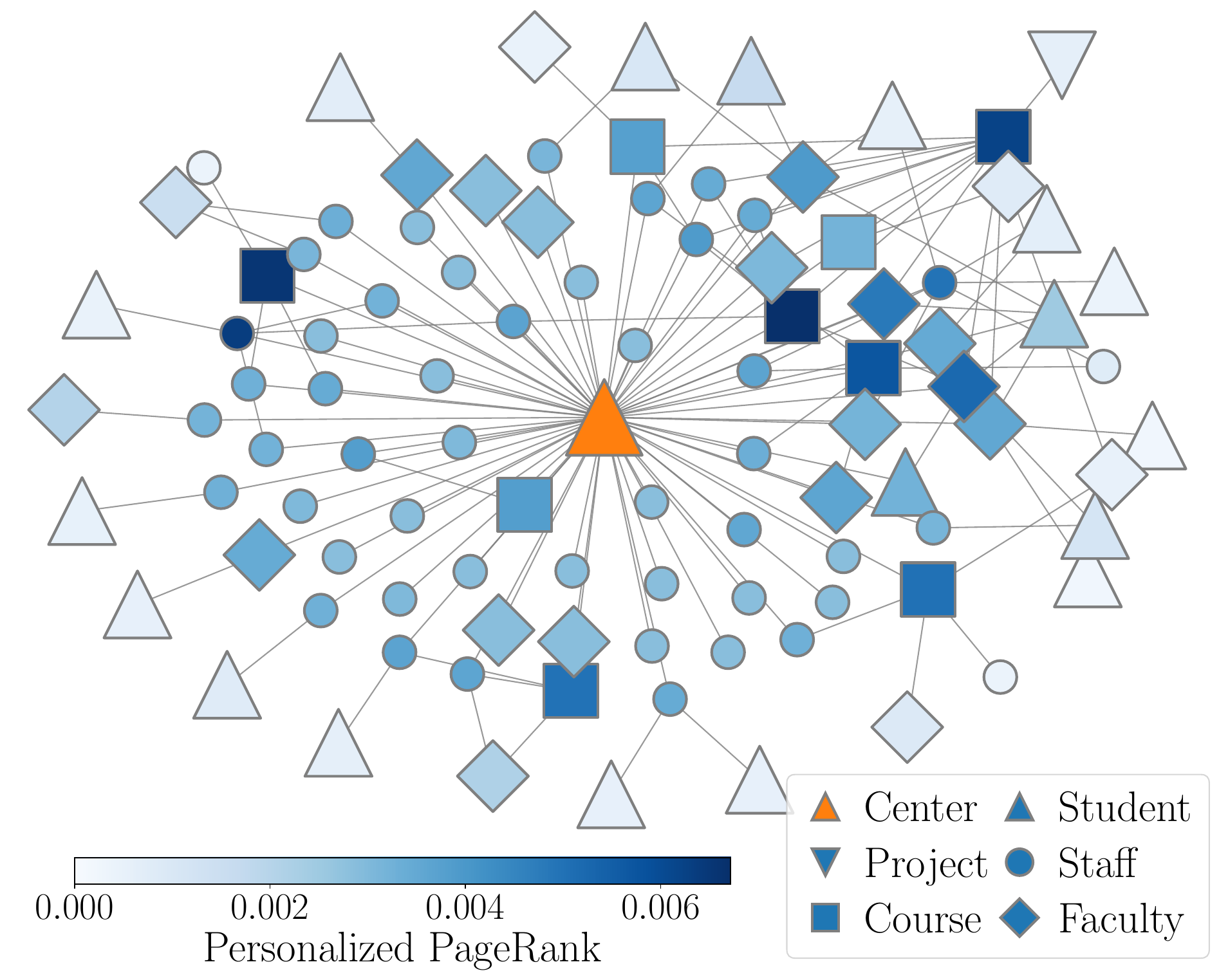}}
    \hfil
    \subcaptionbox{\aggname{} AGG\label{ffig:simrank}}%
    [0.30\linewidth]{\includegraphics[height=1.33in]{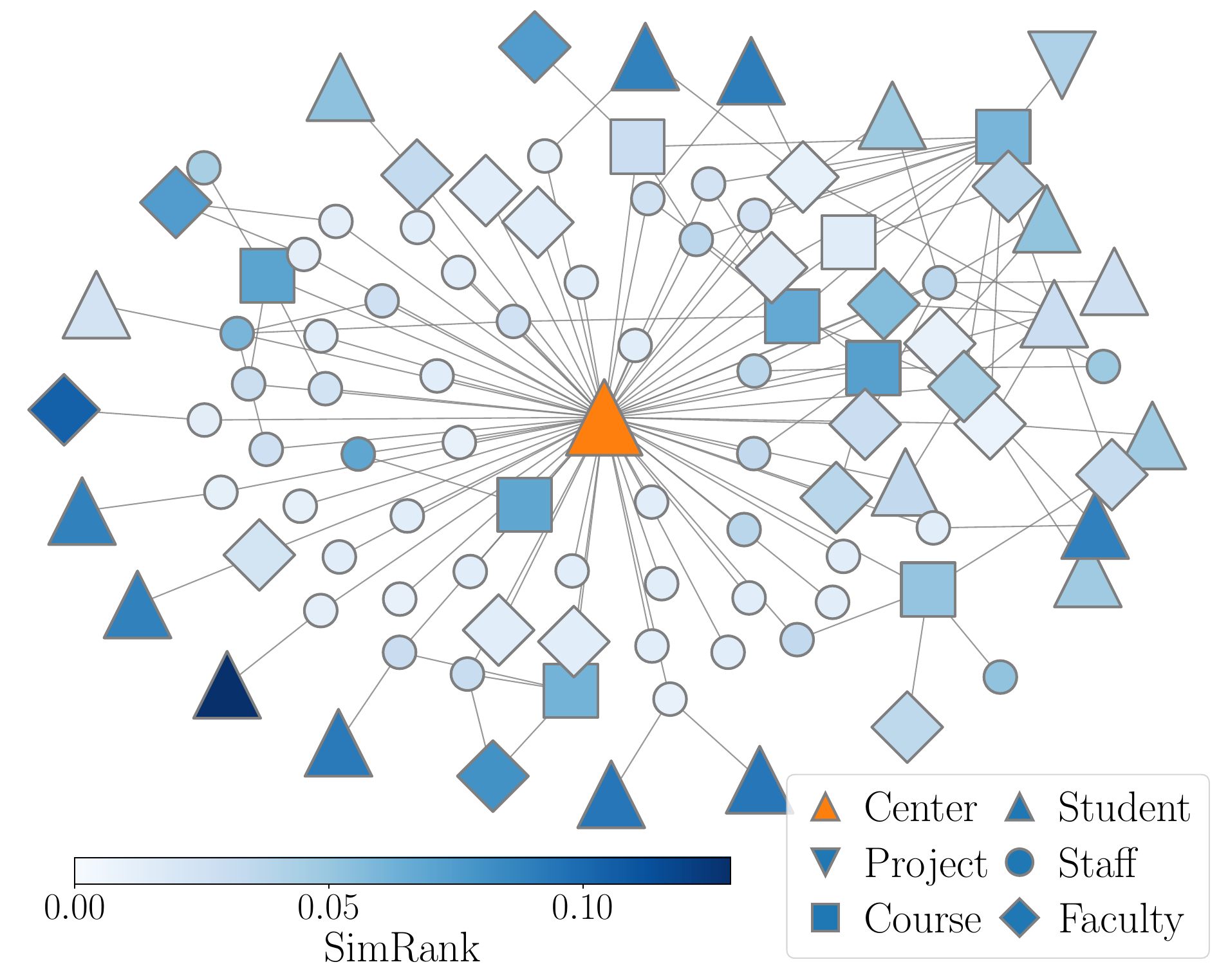}}
\vspace{-0.5ex}
\caption{All sub-figures are from \textit{Texas} heterophily graph. \textbf{(a) A toy example of global structural similarity.} Two staffs inherit high similarity because they share similar neighbors intuitively. \textbf{(\textbf{b}) Neighborhood-based local aggregation and (\textbf{c}) \aggname{} aggregation.} Node color represents aggregation score with respect to the center node (\textcolor{orange}{$\blacktriangle$}). Conventional aggregation focuses on neighboring nodes regardless of node label, while \aggname{} succeeds in assigning high values for nodes with same label (\textcolor{black}{$\blacktriangle$}).}
\label{fig:rank}
\vspace{-0.5ex}
\end{figure*}

Graph neural networks (GNNs) have recently shown remarkable performance in graph learning tasks~\cite{gnn, InductiveRL, yow2022learning, zhu2022spiking, fan2020fusing, agp, gbp}. Despite the wide range of model architectures, traditional GNNs~\cite{gnn} operate under the assumption of \emph{homophily}, which assumes that connected nodes belong to the same class or have similar attributes. In line with this assumption, they employ a uniform message-passing framework to aggregate information from a node's local neighbors and update its representations accordingly~\cite{gnn, gat, appnp}. However, real-world graphs often exhibit \emph{heterophily}, where linked nodes are more likely to have different labels and dissimilar features~\cite{h2gcn}. In these scenarios, the utility of the local and uniform aggregations are limited due to their failure in recognizing distant but similar nodes and assigning distinct attention for nodes of different labels, which causes suboptimal performance under heterophily.

To better apply GNNs in heterophilous graphs, several recent works incorporate long-range relationships and distinguishable aggregation mechanisms. Examples of long-range relationships include the amplified multi-hop neighbors~\cite{mixhop}, and the geometric criteria to discover distant relationships~\cite{geomgcn}. The main challenge in such design lies in deciding and tuning proper neighborhood sizes for different graphs to realize stable performance. 
With regard to distinguishable aggregations,~\citet{pointer} and~\citet{nonlocal} exploited attention mechanism in a whole-graph manner, while~\citet{cosine} and~\citet{glognn} respectively considered feature cosine similarity and global homophily correlation. 
Nonetheless, these approaches require iteratively calculating and updating the correlation for all node pairs, which entails a complexity at least linear to the the number of edges $\mathcal{O}(m)$. When the graph scales up, such aggregation becomes the efficiency bottleneck.

To address the aggregation issues under heterophily, in this paper we propose a novel GNN model with \textbf{Si}mRank-based \textbf{G}NN \textbf{M}essage \textbf{A}ggregation, namely \textbf{\aggname{}}. We highlight two advantages of \aggname{}. Firstly, \aggname{} achieves global and distinguishable aggregation, offering particular adaptability for GNNs under heterophily. Secondly, \aggname{} can be efficiently computed in a one-time manner and necessitates simple aggregation during model updates, which reduces the aggregation complexity to linear to number of nodes $\mathcal{O}(n)$. 

A series of studies have revealed the importance of solely utilizing global and structural information for addressing heterophily in graphs~\cite{geomgcn,large,structure}.~\citet{structure} suggested that nodes surrounded with similar graph structures are likely to share the same label. 
We propose to measure such global similarity in the graph by the SimRank metric, which is based on the intuition that two nodes are similar if they are connected by similar nodes~\cite{simrank}. This can be explained by a toy example in Fig.\ref{ffig:simrank_diag}, that the two nodes representing staffs' websites are more likely to be considered similar and assigned the same label because of their similar neighbors, i.e., respective students. Likewise, student nodes are similar if connecting to similar projects. 
By this means, \aggname{} is able to bypass the dissimilar nodes in the local neighbors, and establish distinct relationships for similar nodes even though they are not directly connected. 
\textcolor{black}{The effectiveness of aggregation on a realistic dataset is shown in Fig.\ref{ffig:pagerank} and ~\ref{ffig:simrank}. In contrast to traditional neighbor-based aggregation for example the PPR, \aggname{} succeeds in discovering distant nodes of the same label as the center node. In fact, we derive a theoretical interpretation in Theorem~\ref{t1} that \aggname{} is capable of capturing global relationships without iterative calculation based on \emph{pairwise} random walk accumulations and prove in Corollary~\ref{cor} that such \emph{pairwise} formation is capable of addressing heterophily issues.} While representing effective global relationships, \aggname{} also enjoys desirable complexity in both precomputation and aggregation. A separated precomputation stage efficiently approximates the SimRank matrix with only $\mathcal{O}(d^2)$ time overhead, where $d=m/n$. After generated, the constant SimRank matrix enables \aggname{} aggregation with $\mathcal{O}(n)$ complexity during GNN training and inference. Compared to previous aggregation schemes of $\mathcal{O}(m)$ or even higher complexities, \aggname{} greatly alleviates the computational bottleneck for performing global aggregation on large-scale graphs. 
We summarize our main contributions as follows:
\begin{itemize}[wide,labelwidth=!,labelindent=0pt,itemsep=0pt,topsep=-1mm]
    \item We propose \aggname{} as a novel GNN model by featuring the node similarity metric SimRank. We theoretically show that \aggname{} is effective in discovering global homophily and grouping similar nodes in heterophilous graphs.
    \item We design a simple and effective scheme for \aggname{} aggregation, which achieves a computational complexity only linear to the number of nodes $\mathcal{O}(n)$ when aggregating global information for all nodes in the graph.
    \item \textcolor{black}{We conduct extensive experiments to showcase the superiority of \aggname{} on a range of datasets with diverse domains, scales, and heterophily properties. Generally, \aggname{} achieves superior performance across 12 datasets and notably, it achieves approximately $4.3\times$ speed-up on average compared to the best baseline, GloGNN in our overall evaluation.}
\end{itemize}

\section{Preliminaries}
\label{sec::preliminaries}
{\color{black}
This section introduces notations and several basic concepts of the Graph Neural Network, Grouping Effect, Graph Heterophily and SimRank algorithm. 

\mypara{Notation.} Table~\ref{tab:fre_notations} shows the notations that are frequently used in this paper. Besides, for a given $\textbf{X}$ matrix, we denote $\textbf{X}(i,j)$ the entry located at $i$-th row, $j$-th column and $\textbf{X}_i$ ($\textbf{X}(:,j)$)  the $i$-th row ($j$-th column) vector.}

\begin{table}[ht]
\centering
\color{black}
\captionsetup{font={color=black}}
\setlength{\abovecaptionskip}{0.5mm}
\setlength{\belowcaptionskip}{0.5mm}
\caption{Frequently used notations in this paper.}
\label{tab:fre_notations}
\renewcommand{\arraystretch}{1.4}
\begin{tabular}{p{.08\textwidth}| p{.32\textwidth}}
\toprule
\textbf{Notations} & \textbf{Descriptions}\\
\midrule
$G = (V,E)$& Undirected graph with node set $V$ and edge set $E$\\\hline
$N_v$ & The neighbors of node $v$\\\hline
$\textbf{A},\textbf{X},\textbf{Y}$ & The input adjacency, node feature and label matrix\\\hline
$\textbf{Z}, \textbf{H}$ & The final updated and middle aggregated node embedding matrix\\\hline
$\textbf{S}, \widehat{\textbf{S}}$ & The original and approximated SimRank matrix \\\hline
$N_y$ & The number of classes\\\hline
$n, m, d$& Numbers of nodes, edges and average degree \\\hline
$k, c$& The top-$k$ schema and decay factor in SIGMA\\\hline
$\epsilon$ & The error threshold parameter \\\hline
$t^{(2\ell)}$ & The tour $t$ of length $2\ell$\\
\bottomrule
\end{tabular}
\end{table}

\mypara{Graph Neural Network.}
Graph neural network is a type of neural networks designed for processing graph data. 
We summarize the key operation in common GNNs in two steps: {\small$\widehat{\textbf{H}}_u^{(\ell)}  = \texttt{AGG}( \{ \textbf{H}_v^{(\ell)}: \forall v \in N_u \} )$} for aggregating neighbor representations and updating {\small$\textbf{H}_u^{(\ell+1)} = \texttt{UPD}(\textbf{H}_{u}^{(\ell)}, \widehat{\textbf{H}}_u^{(\ell)} )$}, where {\small$\textbf{H}_u^{(\ell)}$} denotes the representation of node $u$ in the $\ell$-th network layer, and {\small$\texttt{AGG}(\cdot)$} and {\small$\texttt{UPD}(\cdot)$} are respectively aggregation and update functions specified by concrete GNN models. 
% For a model of $L$ layers in total, the final output of all nodes form the embedding matrix $H^{(\ell)}$, which can be fed to higher-level modules for specific tasks, such as a classifier for node label prediction.
% For example, GCN~\cite{gnn}, representing a series of popular graph convolution networks, uses \texttt{sum} function for neighbor aggregation, and updates the embedding by learnable weights together with \textit{ReLU} activation.
%{Another line of decoupling models~\cite{appnp, wu19sim, liao2022scara} choose to only do aggregation once, hence separating graph propagation and embedding learning. Their updating scheme, or so-called feature transformation, is therefore simple and can be considered as \texttt{MLP} layers learning the embedding matrix.}

\mypara{Grouping Effect.}
Grouping effect~\cite{grouping} describes the global closeness of two nodes in the graph sharing similar features and local structures regardless of their distance. GNN representations exhibiting grouping effect are more effective in node classification under heterophily~\cite{grouping,glognn}. 
For two nodes $u,v \in V$, if (1) their raw feature vectors are similar: $\|\textbf{X}_u-\textbf{X}_v\|_{2}\rightarrow 0$ and (2) neighbors within $L$-hop are similar: $\|\textbf{A}_u^{\ell}-\textbf{A}_v^{\ell}\|_{2}\rightarrow 0$, $\forall \ell \in [1,L]$. The grouping effect of model's output embedding $\textbf{Z}_u$ and $\textbf{Z}_v$ follows $\|\textbf{Z}_u - \textbf{Z}_v\|_2 \rightarrow 0.$

\mypara{Graph Heterophily.}
Graph homophily indicates how similar the nodes to each other with respect to node labels, while heterophily is the opposite~\cite{geomgcn,h2gcn}.
Here we employ node homophily~\cite{geomgcn} defined as the average proportion of the neighbors with the same category of each node:
\begin{equation}
%\scriptsize
\label{eq:h_node}
\begin{matrix}
\mathcal{H}_{node} = \frac{1}{|V|}\sum_{v \in V}\frac{|\{ u \in N_v: y_u =y_v \}|}{|N_v|}.
\end{matrix}
\end{equation}
$\mathcal{H}_{node}$ is in range $(0, 1)$ and homophilous graphs have higher $\mathcal{H}_{node}$ values closer to 1. Generally, high homophily is correlated with low heterophily, and vice versa. 

\mypara{SimRank.}
SimRank is a measure of node pair similarity based on graph topology.
It holds the intuition that two nodes $u,v \in V$ are similar if they are connected by similar neighbors, as calculated by the recursive formula~\cite{simrank} of $\textbf{S}(u,v)=1, u=v$ and for $u\neq v,$ 
\begin{equation}
\label{eq1}
\begin{matrix}
\textbf{S}(u,v) = 
        \frac{c}{|N_u\|N_v|}\sum\limits_{u', v' \in N_u, N_v} \textbf{S}(u',v')
\end{matrix}
\end{equation}
where $c$ is a decay factor empirically set to 0.6. 
Generally, a high SimRank score indicates high structural similarity for a node pair. 

\section{\aggname{} for Heterophily GNN Aggregation}
\label{sec:method}

In this section, we introduce SimRank as a novel GNN model, starting with a pairwise random walk-based interpretation. We then explain how this pairwise approach addresses heterophily issues. Next, we present \aggname{} for GNN message aggregation and prove its grouping effect. Finally, we highlight its superior scalability compared to existing methods.

\subsection{Interpreting SimRank for heterophily}
\label{sec:method:interpreting}

In graphs characterized by node heterophily, neighboring nodes often belong to different categories and exhibit diverse feature distributions. This diversity can hinder the model's ability to aggregate neighbor information meaningfully for predictive tasks such as node classification~\cite{h2gcn}. Consequently, traditional local uniform GNN aggregation yields suboptimal results by inadequately smoothing local dissimilarities, failing to identify intra-class node pairs that are geographically distant~\cite{yan2021two,yang2022}. In contrast, as derived from Eq.~(\ref{eq1}), SimRank is defined for any node pair within the graph and leverages whole-graph structural information. Even over long distances, SimRank effectively assigns higher scores to structurally similar node pairs, thereby extracting relationships on a global scale. \textcolor{black}{Such global property inherently brought by SimRank makes it of great potential in handling heterophily}. Before diving into the main methods, we first demonstrate that \textit{utilizing SimRank to aggregate features implicates global information, which is quite beneficial for heterophily graph learning}. This is based on the concept of \emph{Pairwise} Random Walk, formally defined below:

\begin{definition}[\emph{Pairwise} Random Walk~\cite{simrank}]
\label{thm:def_prw}
The pairwise random walk measures the probability that two random walks starting from node $u$ and $v$ simultaneously and meet at the same node $w$ for the first time. The probability of such random-walk pairs for all tours $t^{(2\ell)}$ with length $2\ell$ can be formulated as:
\begin{align*}
%\label{eq10}
    \overleftrightarrow{P}(u, v|t^{(2\ell)}) &= \begin{matrix}
    \sum_{t^{(2\ell)}}p(x|u,t^{(\ell)}_{u:x})\cdot p(x|v, t^{(\ell)}_{v:x})
    \end{matrix} \\
    &= \begin{matrix}\sum_{w \in V} p(w|u, t_{u:w}^{(\ell)}) \cdot p(w|v, t_{v:w}^{(\ell)})\end{matrix},
\end{align*}
where {$t_{u:v}^{(2\ell)}$} is one possible tour of length {$2\ell$, $t_{v:w}^{(\ell)}: \{v, ..., w\}$} is the sub-tour of length $\ell$ compositing the total tour, and {$p(w|v, t^{(\ell)}_{v:w})$} denotes the probability a random walk starting from node $v$ and reaching at node $w$ under the tour {$t^{(\ell)}_{v:w}$.} 
\end{definition}
Since random walk visits the neighbors of the ego node with equal probability, {$p(w|v,t^{(\ell)}_{v:w})$} can be calculated as {$p(w|v,t^{(\ell)}_{v:w})=\prod_{w_i\in t_{v:w}}\frac{1}{|N_{w_i}|}$}, where $w_i$ is the $i$-th node in tour {$t^{(\ell)}_{v:w}$}. Generally, a higher probability of such walks indicates strong connectivity between the source and end nodes. {\color{black}Note that interpreting SimRank through random walk probabilities, such as the meeting probability of two $\sqrt{c}$-walks, is common in its efficient computation~\cite{wang2020exact}. Here, we leverage SimRank with the pairwise random walk to provide greater intuition for addressing graph heterophily in the subsequent theorems.} Pairwise random walk paves a way for node pair aggregation that can be calculated globally, compared to conventional GNN aggregation that is interpreted as random walks with limited hops, which is highly local~\cite{appnp,Xu2018Representation}. In fact, we then show that SimRank based aggregation can be decomposed into form of such walk probabilities and thus inherently shares the global property in the following theorem. 

\begin{theorem}
\label{t1}
    On graph $G$ with SimRank matrix $\textbf{S}$ and arbitrary initialized node embedding matrix $\textbf{H}$, denoted the SimRank aggregated feature matrix as $\widehat{\textbf{Z}}=\textbf{SH}$, for each node $u \in V$, we have:
\begin{align}
\label{eq_agg}
\begin{matrix}    
\widehat{\textbf{Z}}_u=\sum_{\ell=1}^{\infty} c^{\ell}\sum_{v\in V} \overleftrightarrow{P}(u, v|t^{(2\ell)}) \cdot \textbf{H}_v.
\end{matrix}
\end{align}
\end{theorem}

\begin{proof}
We first suppose for any node pair $(u,v)$, it holds that 
\begin{align*}
\nonumber
\textbf{S}(u,v) %&= \sum_{\ell=1}^{\infty} c^{\ell} \cdot  \overleftrightarrow{P}(u, v|t^{(2\ell)}) \\
&= \sum_{\ell=1}^{\infty} c^{\ell}\sum_{w \in V} p(w | u, t_{u:w}^{(\ell)}) \cdot p(w | v, t_{v:w}^{(\ell)}).
\end{align*}

{\color{black}
Denoted {\small$\textbf{S}'(u, v) = \sum_{\ell=1}^{\infty} c^{\ell} \cdot  \overleftrightarrow{P}(u, v|t^{(2\ell)})$}, it can be easily verified that {${\small\textbf{S'}}(u,v)=1$} if $u=v$, and {${\small\textbf{S}'}(u,v)=0$} if there is no tour $t$ consisting of two separate random walk paths starting from $u$ and $v$ and ends in any same node $w$. Then, by performing one step random walk, we split {${\small\textbf{S}}'(u,v)$} as
{\footnotesize\begin{align*}
    &=  \sum_{\ell'\geq 0, } \; \sum_{u', v'} c^{(\ell'+1)} \cdot P(u|u',t^{(1)}) \cdot P(v|v',t^{(1)}) \cdot  \overleftrightarrow{P}{(u',v'|t^{\ell'})} \\
    &= \sum_{\ell'\geq 0}\; \sum_{u', v'} \frac{c^{(\ell'+1)}}{|N_u\|N_v|} \cdot  \overleftrightarrow{P}{(u',v'|t^{\ell'})}     %&\quad\ (\textbf{by } \overleftrightarrow{P}(u',v'|t^{(0)})=0, u\neq v) \\
    = \frac{c}{|N_u\|N_v|} \sum_{u', v' } \sum_{\ell'\geq 0} c^{\ell'} \overleftrightarrow{P}{(u',v'|t^{\ell'})}  \\
    &= \frac{c}{|N_u\|N_v|} \sum_{u', v' } \sum_{\ell'\geq 1}  c^{\ell'} \overleftrightarrow{P}{(u',v'|t^{\ell'})} = \frac{c}{|N_u\|N_v|} \sum_{u', v' } \textbf{S}'(u',v').
\end{align*}}
Here, the last term is exactly identical to the original SimRank definition Eq.~(\ref{eq1}). Due to the uniqueness of the solution to Eq.~(\ref{eq1}), we conclude that {\small$\textbf{S}'(u,v)=\textbf{S}(u,v)$}. 
Substituting it into calculation of $\textbf{SH}$, we get:
\begin{align}
\nonumber
\begin{matrix}
\widehat{\textbf{Z}}_u = \sum_{v\in V} \textbf{S}(u, v) \cdot \textbf{H}_v = \sum_{v\in V} \sum_{\ell=1}^{\infty} c^{\ell} \overleftrightarrow{P}(u, v|t^{(2\ell)}) \cdot \textbf{H}_v,
\end{matrix}
\end{align}
which completes the proof.}
\end{proof}
The implication above is by two folds. {\color{black} Firstly, SimRank based aggregation is \textit{global}, as ${\small\textbf{S}}(u, v)$ accumulates the pairwise random probabilities of {\small$\overleftrightarrow{P}(u, v|t^{(2\ell)})$}, which counts on all possible tours connecting the node pair $(u,v)$ to assign useful weights for node features, no matter how distant the nodes $u$ and $v$ are in the graph. This shows greater global property compared to traditional local aggregation based on single path random walk probabilities.}
Secondly, it naturally retrieves such global relationship in an \textit{one-time} manner, while the neighbor-based representation requires sufficient number layers conducting local aggregations to achieve global view. Based on the above implications, we further develop Corollary~\ref{cor} to demonstrate that non-vanishing pairwise random scores {\small$\overleftrightarrow{P}(u, v|t^{(2\ell)})$} will grasp homophily node pairs of $u,v$ and link such property to the overall performance more directly.

\begin{corollary}
\label{cor}
    The aggregated node \texttt{$v$} (with non-vanishing aggregation weights) with respect to each central node \texttt{$u$} based on Eq.(\ref{eq_agg}) is more likely to be \emph{homophily} with the graph heterophily extent grows. 
\end{corollary}
\begin{proof}
Recall that in Theorem~\ref{t1}, the updated embedding $\widehat{\textbf{Z}}_u$ for the central node $u$ aggregates nodes with non-zero pairwise random meeting probabilities. Despite its global capability, each tour $t$ of length 
$2\ell$ is of \emph{even-hop}, which is inherently linked to the homophily of the message-passed node sets. Specifically, we here consider binary classification tasks and assume the probability of a neighbor holding a different label than its central node is $p$ (representing the heterophily extent). {\color{black} Upon message passing from node $u$ to node $v$ through $t^{(2\ell)}$, we first calculate the probability of nodes $u$ and $v$ being homophilic denoted as $H_p^{\ell}$. Based on the tour $t^{(2\ell)}:\{u,u'...v',v\}$, we can conduct the calculated of $H_p^{\ell}$ by recursion:
$$
H^\ell_p=p^2 \cdot H^{\ell-1}_p + (1-p)^2 \cdot H^{\ell-1}_p,
$$
where the $H^{\ell-1}_p$ means node $u'$ and $v'$ are being homophilic under path $t^{(2\ell-2)}$. Putting one step forward, nodes $u$ and $v$ are homophilic when $u,u'$ and $v,v'$ are all homophilic, with probability $(1-p)^2$ or $u,u'$ and $v,v'$ are of different class with probability $p^2$. when $\ell=0$, we have $H^0_p=1$ and thus we can get the result as:
\begin{align}
\nonumber H^\ell_p&=p^2 \cdot H^{\ell-1}_p + (1-p)^2 \cdot H^{\ell-1}_p\\
\nonumber&=(2p^2-2p+1)\cdot H^{\ell-1}_p=...=(2p^2-2p+1)^\ell.
\end{align}
Next, we conduct step-by-step derivation of {\small\(\tfrac{d}{dp}H_p^\ell\)}.
Denote $f(p)=(2p^2-2p+1)$ and then {\small$H_p^\ell = \bigl[f(p)\bigr]^\ell$}. Differentiate w.r.t. $p$ using the chain rule, we get $\frac{d}{dp}\Bigl[f(p)^\ell\Bigr] =\ell \bigl[f(p)\bigr]^{\ell - 1}\cdot\frac{d}{dp}\bigl[f(p)\bigr].$ Then we compute \(\frac{d}{dp}f(p)\) as 
$$f'(p) \;=\; \frac{d}{dp}\bigl(2p^2 - 2p + 1\bigr)= 2(p-2).$$
Putting these steps together we get:
\[
\frac{d}{dp}\Bigl[(2p^2 - 2p + 1)^\ell\Bigr]
\;=\;
\ell \,\bigl(2p^2 - 2p + 1\bigr)^{\ell-1} 
\;\cdot\;
(4p - 2).
\]
Hence, the derivative is derived as follow:
\[
\boxed{
\frac{d}{dp}H_p^\ell 
=
2\ell \cdot \bigl(2p - 1\bigr)
\;\bigl(2p^2 - 2p + 1\bigr)^{\ell-1}
\;.
}
\]
Under heterophily graph settings with $p>0.5$ such that $\Delta > 0$, we'll observe two key implications as the heterophily extent $p$ grows: (1) Larger $\ell$ results in a faster increase in the homophily probability and (2) $\Delta$ increases rapidly as $p$ grows, causing $H_p^{\ell}$ to approach one quickly. In other words, \emph{the probability of a target node being homophilic increases as the graph becomes more heterophilic and the heterophily extent increases}. Additionally, implication (1) underscores the importance of long-range connections particularly in current tour structures, which further highlights the importance of introducing the global aggregation of $\textbf{S}$ in Theorem~\ref{t1}.}
\end{proof}

\begin{figure}[!t]
{
\vspace{-0.5ex}
\includegraphics[width=0.48\textwidth]
{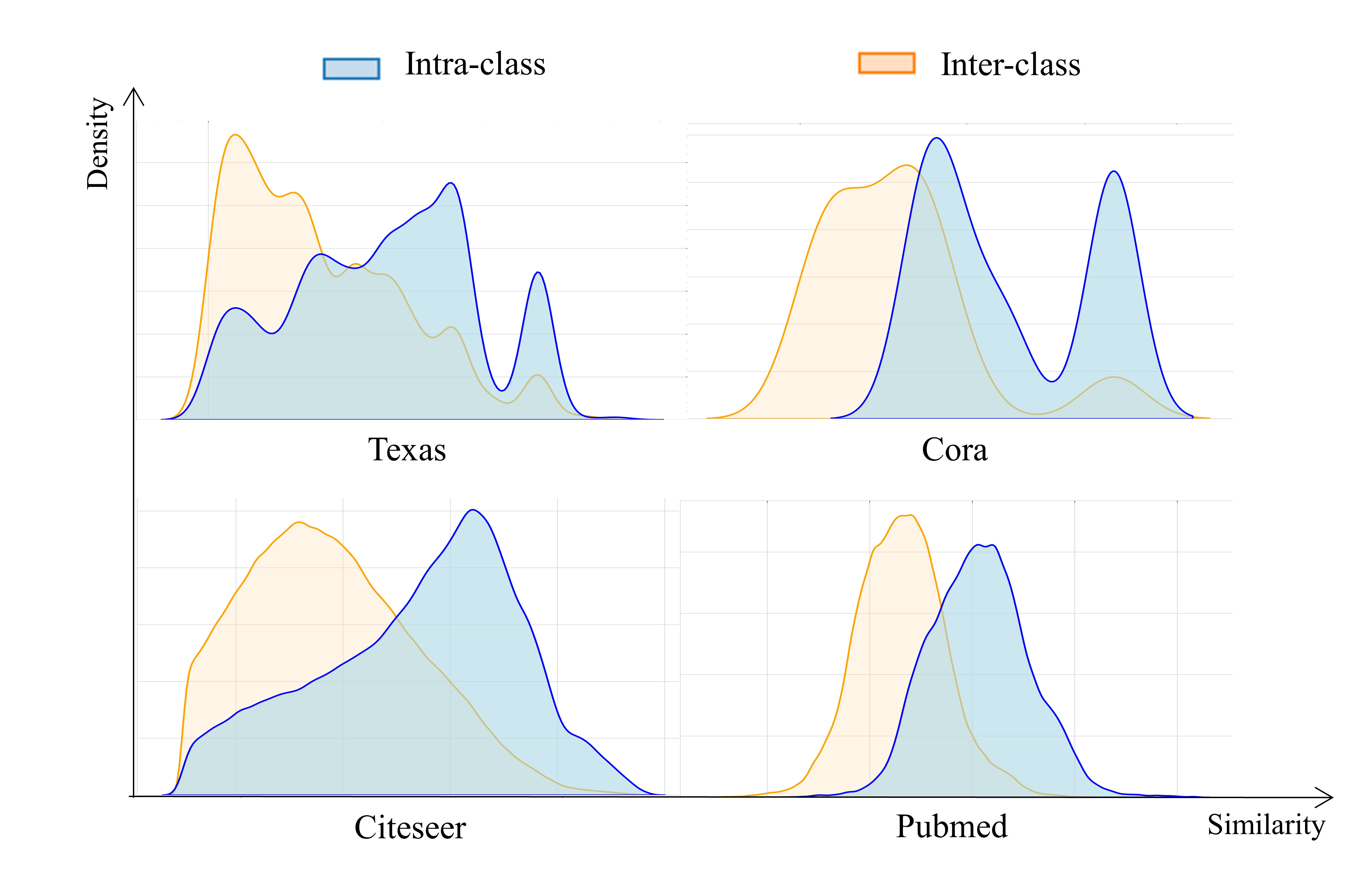}}  
\vspace{-0.5ex}
\captionsetup{width=0.48\textwidth}
\caption{Patternes of SimRank scores over intra-class and inter-class node pairs. X-axis denotes the similarity score corresponding to one node pair and Y-axis the density.} 
\label{distribution}
\vspace{-0.5ex}
\end{figure}

In fact, the above theorems summarize the advantage of SimRank in addressing heterophily issues. By accumulating pairwise random walk probabilities, SimRank effectively identifies homophilous nodes for each central node by considering a global range of nodes. In other words, nodes assigned higher SimRank scores relative to a central node are more likely to share homophilous properties. We further validate this property using several representative real-world heterophilous graphs. As shown in Fig.\ref{distribution}, the densities of SimRank scores for node pairs across four different graphs are displayed. The distribution of similarity scores distinctly varies between intra- and inter-class node pairs, indicating a clear pattern of score allocation based on label homophily.  Additionally, a closer analysis of the mean and standard deviation values in Table~\ref{tab:statistics} reveals that intra-class node pairs generally achieve higher SimRank scores compared to inter-class pairs within the same graph. For example, in the \emph{Texas} dataset, intra-class pairs exhibit a mean SimRank score of 0.21, compared to 0.16 for inter-class pairs. This robust alignment of SimRank with homophily suggests its effectiveness in identifying homophily relationships. We emphasize that this characteristic significantly enhances the aggregation process by prioritizing representations from homophily nodes for each central node, thereby improving the model's learning ability in heterophily graphs. Combining these theoretical insights with practical evidence, we identify SimRank as a global similarity measure particularly well-suited for graphs under heterophily, which serves as a key inspiration for the design of our \aggname{} architecture, as detailed in the following part.

\begin{table}[!t]
\centering
\renewcommand{\arraystretch}{1.5}
\captionsetup{width=0.48\textwidth}
\caption{Mean $\&$ standard variance of node-pair similarities.}
\vspace{-0.5ex}
\resizebox{0.48\textwidth}{!}{
\begin{tabular}{c|cccc}
\toprule
\textbf{Type} & \textbf{Texas} & \textbf{Chameleon} & \textbf{Cora} & \textbf{Pubmed} \\ 
\midrule
\textbf{Intra-class} &	0.21±0.07	& 0.16±0.03	& 0.35±0.06	&0.13±0.02 \\ \hline
\textbf{Inter-class}&	0.16±0.06&	0.13±0.03&	0.31±0.06&	0.11±0.01 \\
\bottomrule
\end{tabular}}
\label{tab:statistics}
\vspace{-0.5ex}
\end{table}

\subsection{\aggname{} Aggregation Workflow} 
\label{sec:method:grouping}

As SimRank assigns high scores to node relations with topological similarities, it can be regarded as a powerful global aggregation for GNN representation, guiding the network to put higher weights on such distant but homophilous node pairs. Hence, our goal is to design the effective and efficient GNN model \aggname{} that exploits the desirable capability of SimRank. Based on the implication of Theorem~\ref{t1} that SimRank already carries global similarity information implicitly, our model removes the need of iteratively aggregating and updating embeddings by incorporating the aggregation matrix such as~\cite{cosine,glognn}. Instead, we only rely on a constant SimRank matrix $\textbf{S}$ which can be efficiently calculated in precomputation. To generate node representations from graph topology and node attributes, we deploy a simple and effective heterophilous GNN architecture, derived from LINKX~\cite{large} that respectively embeds the adjacency matrix $A$ and attribute matrix $X$ by two \texttt{MLP}s, i.e., $\texttt{MLP}_A$ and $\texttt{MLP}_X$, then joins by a third $\texttt{MLP}_H$. A tunable parameter $\delta \in [0, 1]$, namely the feature factor, is employed to control the combination. 
The node representation matrix $H$ is:
\begin{align}
\begin{split}
    \label{eq15}
    \mathbf{H}& \mathbf{_A} = \texttt{MLP}_A(\textbf{A}),\; \mathbf{H_X} = \texttt{MLP}_X(\textbf{X}), \\
    \textbf{H} &= \texttt{MLP}_H \left(\delta \cdot \mathbf{H_X} + (1-\delta) \cdot \mathbf{H_A} \right). 
\end{split}
\end{align}

\begin{figure}[!t]
\centering
\includegraphics[width=0.36\textwidth]{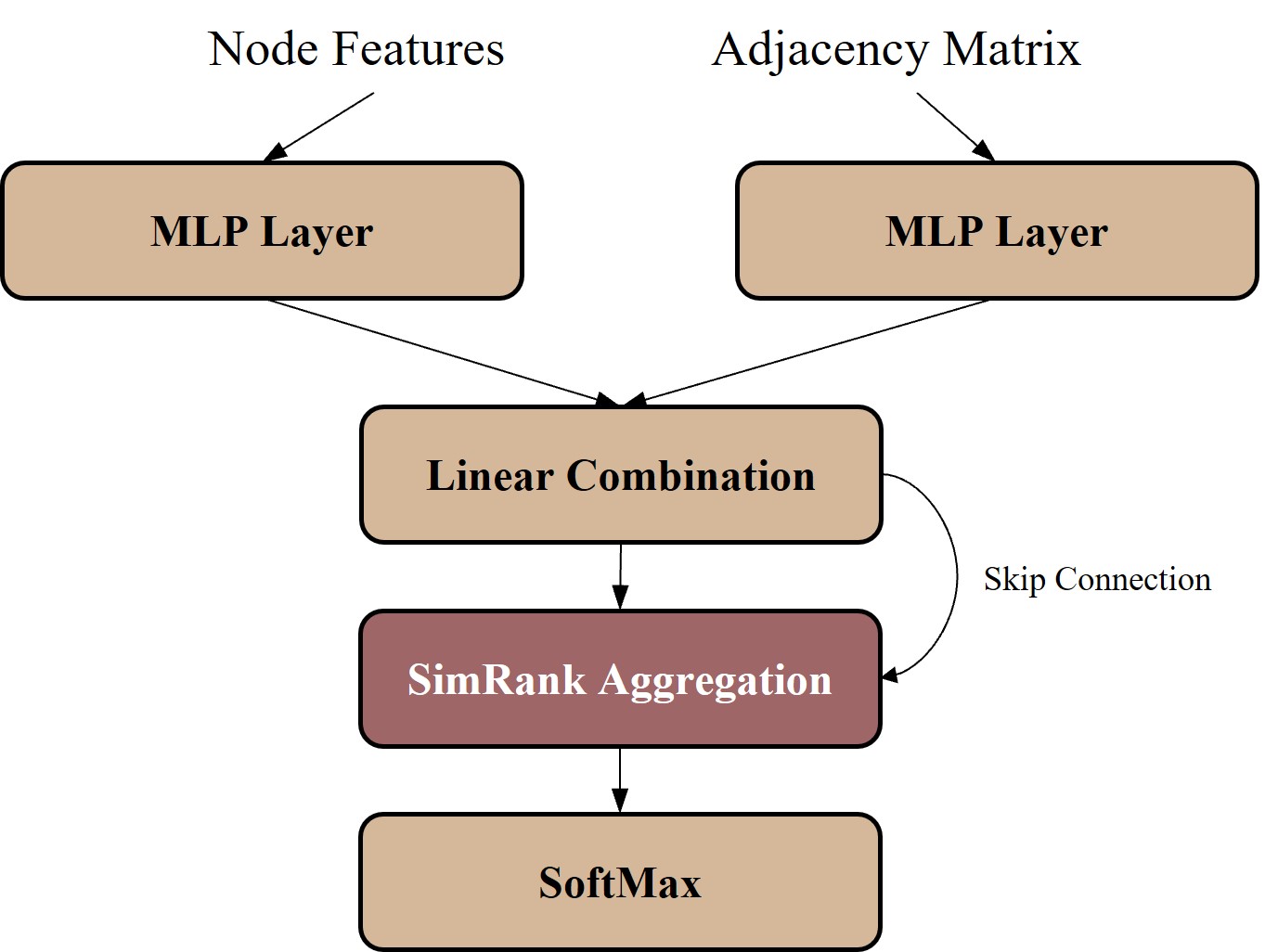}
  \caption{Architecture of \aggname{}.}
  \label{architecture}
  \vspace{-1.ex}
\end{figure}

For each node $u\in V$, based on the joint node representation vector $h_u$, we are able to apply the matrix $\textbf{S}$ to gather similar nodes globally by corresponding SimRank values. Our \aggname{} aggregation and update can be respectively described as:
\begin{align}
    \label{eq:zsim}
    \texttt{AGG}:\; \widehat{\textbf{Z}}_u &= \begin{matrix}
        \sum_{v\in V}
    \end{matrix} \textbf{S}(u, v) \cdot \textbf{H}_v, \\
    \label{eq8}
    \texttt{UPD}:\; \mathbf{Z_u} &= (1 - \alpha)\cdot\widehat{\textbf{Z}}_u + \alpha \cdot \textbf{H}_u,
\end{align}
\noindent where parameter $\alpha \in [0, 1]$ is to balance global aggregation and raw local embeddings. It is notable that instead of only considering neighboring nodes $v\in N_u$, Eq.~(\ref{eq:zsim}) aggregates all potential nodes in the graph and differentiates the importance by mutual similarity. Such aggregation scheme thus bypasses the potential negative influence of heterophilous nodes in the neighborhood. With the topology and attribute input $A$ and $X$, we validate the grouping effect of \aggname{} below:
\begin{theorem}
\label{t2}
Node representation $\textbf{Z}$ has grouping effect when the hop condition $L \geq log_c{\epsilon}$, where $\epsilon$ is the absolute error requirement in matrix $\textbf{S}$ approximating calculation. 
\end{theorem}
\begin{proof}
For nodes $u, v$ defined in grouping
effect conditions: {$$\|\textbf{X}_u-\textbf{X}_v\|_{2}\rightarrow 0, \text{ and } \|\textbf{A}_u^{\ell}-\textbf{A}_v^{\ell}\|_{2}\rightarrow 0, \forall \ell \in [1,L]. $$} Denote {$\widehat{\textbf{H}}=\delta \cdot \mathbf{H_X} + (1-\delta) \cdot \mathbf{H_A}$}, then for {$\forall i \in \{1, ..., f\}$}, 
\begin{align*}
\text{we have } &\widehat{\textbf{H}}(u,i)-\widehat{\textbf{H}}(v,i)\\
&=\delta \cdot \left[ \mathbf{H_X}(u,w)-\mathbf{H_X}(v,w) \right] +\\
&\quad (1-\delta) \cdot \left[\mathbf{H_A}(u,i) - \mathbf{H_A}(v,i)\right] \\
&\leq \begin{matrix}\sum_{j\in V} (1-\delta)\cdot|\mathbf{A}(u,k)-\mathbf{A}(v,k)|\cdot \textbf{W}(k,i)\end{matrix} +\\
&\quad  \delta \cdot |\textbf{X}(u,j)-\textbf{X}(v,j)|\cdot \textbf{W(}k,i),
\end{align*}
where $\textbf{W}$ denotes the $\texttt{MLP}$ weights. Based on this, we have 
\begin{align*}
\nonumber
&\|\widehat{\textbf{H}}(u,i)-\widehat{\textbf{H}}(v,i)\|_2\leq \delta^2 \cdot \|\textbf{W}{(:,i)}\|_2 \cdot \|\textbf{X}_u - \textbf{X}_v\|_2+\\
& (1-\delta)^2 \cdot \|\textbf{W}(:,i)\|_2 \cdot \|\textbf{A}_u - \textbf{A}_v\|_2 \rightarrow 0.
\end{align*}
Hence, we've concluded that the matrix $\widehat{\textbf{H}}$ holds grouping effect. Due to the continuous properties of the $\texttt{MLP}$ functions, we can infer that matrix $\textbf{H}=\texttt{MLP}_H(\widehat{\textbf{H}})$ also shares such property. Next we consider the aggregation matrix $\textbf{S}$. According to~\cite{top-k-simrank}, its $\epsilon$-approximation can be achieved through $T$ iterations of matrix multiplication, where for $\forall\  u,v, |\textbf{S}(u,v) - \textbf{S}^T(u,v)| < \epsilon$, we can get $\textbf{S}^T = \begin{matrix}\sum_{\ell=0}^{\lceil \log_c\epsilon \rceil}\;\end{matrix}c^\ell (1-c) (\textbf{P}^\ell)(\textbf{P}^\top)^\ell$. Then we first conclude that {$\forall \ell \in [1,L], \|\textbf{P}_u^\ell-\textbf{P}_v^\ell\|_{2}\rightarrow 0$} holds by math induction. when $\ell=1$, both $\textbf{P}=\textbf{D}^{-1}\textbf{A}$ and $\textbf{P}^\top=\textbf{AD}^{-1}$ hold since $\textbf{D}$ is the row sums of $\textbf{A}$. Suppose {$\|\textbf{P}_u^{\ell-1}-\textbf{P}_v^{\ell-1}\|_{2}\rightarrow 0$}, we have that for $\forall i, \|\textbf{P}^\ell(u,i)-\textbf{P}^\ell(v,i)\|_2$
\begin{equation}
\nonumber
\begin{split}
& = \|\textbf{\textbf{P}}^{\ell-1}(u,:) \textbf{P}(:,i) - \textbf{P}^{\ell-1}(v,:) \textbf{P}(:,i)\|_2 \\
&\leq \|\textbf{P}^\top_i\|_{2} \cdot \|\textbf{P}_u^{\ell-1}-\textbf{P}_v^{\ell-1}\|_{2} \rightarrow 0 \Rightarrow \|\textbf{P}_u^\ell-\textbf{P}_v^\ell\|_{2}\rightarrow 0.
\end{split}
\end{equation}
%Since for \haoyu{math induction for $P^\ell$} {$\forall \ell \in [1,L], P^\ell=(D^{-1}A)^\ell.$} We further derive the $u$-th row of {$P^\ell$} as { $P_u^\ell = \sum_{w \in V} a_u^1[w] \cdot a_u^\ell = d_u^\ell \cdot a_u^\ell.$
%When $\|a_u^\ell-a_v^\ell\|_{2}\rightarrow 0$}, we have {$\|d_u^\ell-d_v^\ell\|_{2}\rightarrow 0$} and consequently {$\|P_u^\ell-P_v^\ell\|_{2}\rightarrow 0$} holds. 
Based on this, we then show that matrix $\textbf{S}^{*}=\textbf{S}-(1-c)\cdot \textbf{I}$ holds the grouping effect as for {$\forall i, \textbf{S}^*(u,i) - \textbf{S}^*(v, i)$}
\begin{align*}
\nonumber
&=\begin{matrix} \sum_{\ell=1}^{\lceil \log_c\epsilon \rceil} \end{matrix} c^\ell (1-c) \left[ \textbf{P}_u^\ell (\textbf{P}_i^\ell)^\top - \textbf{P}_v^\ell (\textbf{P}_i^\ell)^\top\right]\\ 
&=\begin{matrix}\sum_{\ell=1}^{\lceil \log_c\epsilon \rceil} \end{matrix} \left[ \textbf{P}_u^\ell - \textbf{P}_v^\ell \right] \left[ c^\ell (1-c) (\textbf{P}_i^\ell)^\top \right].
\end{align*}
With {$L\geq\log_c\epsilon$}, We then have that for {$\|\textbf{S}^*(u,i) - \textbf{S}^*(v, i)\|_{2}$}
\begin{align*}
\nonumber
&\leq \begin{matrix}
\sum_{\ell=1}^{\lceil \log_c\epsilon \rceil}\end{matrix}
\|\textbf{P}_u^\ell - \textbf{P}_v^\ell\|_{2} \cdot \| c^\ell (1-c) (\textbf{P}_i^\ell)^\top \|_2 \\
&\leq \lceil \log_c\epsilon \rceil \cdot \|c(1-c)\cdot \textbf{P}_i^1\|_{2} \cdot \|\textbf{P}_u^\ell - \textbf{P}_v^\ell\|_{2} \rightarrow 0.
\end{align*}
Therefore, we conclude that matrix $\textbf{S}^*$ hold the expected grouping effect. We then transform Eq.~(\ref{eq:zsim}) and ~(\ref{eq8}) into
\begin{align*}
\centering
\nonumber
&\textbf{Z}=\alpha \cdot \textbf{S}^* \cdot \textbf{H} + (1-\alpha) \cdot c \cdot \textbf{H} \\
&\Rightarrow  \forall i, \|\textbf{Z}(u,i)-\textbf{Z}(v,i)\|_{2}=  
\|\alpha \cdot [\textbf{S}^*_u-\textbf{S}^*_v] \cdot \textbf{H}(:,i) + \\
&(1-\alpha) \cdot c \cdot [\textbf{H}(u,i)-\textbf{H}(v,i)]\|_{2}  \leq 2\alpha\cdot \|\textbf{H}(:,i)\|_{2}\cdot \\
&\|\textbf{S}^*_u-\textbf{S}^*_v\|_{2} + 4c^2\cdot (1-\alpha)^2 \cdot \|\textbf{H}(u,i)-\textbf{H}(v,i)\|_{2}.
\end{align*}
Since both matrices $\textbf{S}*$ and $\textbf{H}$ have grouping effect, we have
\begin{equation}
\nonumber
\|\textbf{S}^*_u-\textbf{S}^*_v\|_{2} \rightarrow 0 \text{ and } \|\textbf{H}(u,i)-\textbf{H}(v,i)\|_{2} \rightarrow 0,
\end{equation}
which contributes to the conclusion that {$ \|\textbf{Z}_u-\textbf{Z}_v\|_{2} \rightarrow 0$.} Till now, we've verified that matrices $\textbf{H}$, $\textbf{Z}$ and $\textbf{S}^*$ all have the desired grouping effect, which completes the proof.
\end{proof}

The desired grouping effect of $\textbf{Z}$ demonstrates the effectiveness of \aggname{}, in that for two nodes, regardless of their distance, if they share similar features and graph structures, their aggregated representations will be similar and are more likely to be classified under the same label. On the contrary, for inter-class nodes, due to the low SimRank value $\textbf{S}(u,v)$, their representations will be different. We therefore claim that \aggname{} is capable of handling heterophily graphs, thanks to the distinguishable aggregation procedure. {\color{black}Note that, theoretically, the condition $L \geq \log_c(\epsilon)$ ensures SIGMA's grouping effect. Larger $c$ and smaller $\epsilon$ impose stricter requirements on $L$, meaning a larger $L$ is needed for a real effect. In practice, we commonly choose $c = 0.6$ and $\epsilon = 0.1$, resulting in $L \approx 4$, which is more suitable for real-world datasets. $c = 0.6$ is a standard choice in many SimRank applications~\cite{simpush, simrank, gsim}, and $\epsilon = 0.1$ provides a sufficiently rough approximation to capture structural bias and maintain good efficiency, which will be validated in our experiments section.}

\subsection{Complexity Optimization and \textcolor{black}{Analysis}}
\label{ssec:scale}

{\color{black}We first analyze the computational complexity of SIGMA. Following the procedure in Fig.\ref{architecture}, the first two \texttt{MLP} layers transform the raw features and adjacency matrix, incurring $O(nf^2)$ and $O(ndf) = O(mf)$ costs, respectively. Note that when using \texttt{MLP}($\textbf{A}$) to perform the calculation: $\texttt{MLP}(A) = \sigma(\textbf{AW})$ in PyTorch, the products of $\textbf{AW}$ are handled with sparse-dense multiplication without ever densifying $A$. As such, the overall memory and computational overhead is $\mathcal{O}(m + nf)$ and $\mathcal{O}(mf)$, respectively. This implementation is applicable to large-scale graphs with costs linear in the number of edges. 

The most time-consuming part is the SimRank aggregation as $\widehat{\textbf{Z}}_u = \sum_{v \in V} \textbf{S}(u, v) \cdot \textbf{H}_v$, which incurs $O(n^2f)$ (if using the full dense SimRank matrix of $\mathcal{R}^{n \times n}$) in computation and outputs logit vectors of shape $\mathbb{R}^{n \times N_y}$. Each node vector is then passed through a SoftMax function to produce class probabilities, contributing an overhead of $O(nN_y)$ for all nodes. Since $f$ and $N_y$ are relatively small constants, SoftMax's impact on runtime is negligible.} {As such, the aggregation step dominates the calculation time. To address the computation bottleneck of global aggregation in SIGMA, we first decouple the SimRank calculation into an individual precomputation stage.} By the definition in Eq.~(\ref{eq1}), if the full and exact SimRank matrix $\textbf{S} \in \mathcal{R}^{n \times n}$ is utilized, the recursive computation will demand a time complexity of $\mathcal{O}(n^2d^2)$, which is not scalable, especially when the graph size $n$ is large. The $\mathcal{O}(n^2)$ dense matrix also makes the Eq.~(\ref{eq:zsim}) aggregation impractical. To address this computation complexity, we choose to utilize an approximation of the SimRank scores, benefiting from a range of efficient and scalable algorithms~\cite{simpush,luo2023massively}. We apply Algorithm~\ref{simpush} (LocalPush) as a precomputation to acquire the SimRank approximation. This computation efficiently estimates significant SimRank scores within a precision guarantee $\epsilon$. Its time complexity can be bounded by the following lemma:

\begin{lemma}[\cite{simpush}] 
    % Input $P=0$ and $R=I$, 
    Algorithm 1 returns an approximate matrix $\widehat{\textbf{S}}$ in $\mathcal{O}(\frac{d^2}{c(1-c)^2\epsilon})$ time with $\|\widehat{\textbf{S}}-\textbf{S}\|_{max} < \epsilon$.
\end{lemma}

\setlength{\textfloatsep}{0pt}

\begin{algorithm}[!b]
\algrenewcommand{\alglinenumber}[1]{\scriptsize\bfseries#1}
\renewcommand{\algorithmicrequire}{\textbf{Input:}}
\renewcommand{\algorithmicensure}{\textbf{Output:}}
\caption{Localpush~\cite{simpush}}
\label{simpush}
\begin{algorithmic}[1]
    \Require Graph $G$, decay factor $c$, error threshold $\epsilon$ 
    \Ensure Approximate SimRank matrix $\widehat{\textbf{S}}$
    \State $\widehat{\textbf{S}} \gets 0,\, \textbf{R} \gets \textbf{I}$
    \While{max$_{(u,v)}$ $\textbf{R}(u,v) > (1-c)\epsilon$}
        \State $\widehat{\textbf{S}}(u,v) \leftarrow \widehat{\textbf{S}}(u,v)+\textbf{R}(u,v)$
        \ForAll{$u' \in N_u, v' \in N_v$}
            \State $\textbf{R}(u',v') \leftarrow \textbf{R}(u', v') + c\cdot \frac{\textbf{R}(u,v)}{|N_{u'}|\cdot |N_{v'}|}$
        \EndFor
        \State $\textbf{R}(u,v) \leftarrow 0$
    \EndWhile
    {\color{black}$\widehat{\textbf{S}}(u,v) \leftarrow 0$, \text{if } $\widehat{\textbf{S}}(u,v) < \frac{1}{10} \epsilon$ \texttt{// prune trivial.}}
    \State \Return matrix $\widehat{\textbf{S}}$

\end{algorithmic}
\end{algorithm}
\setlength{\textfloatsep}{6pt plus 2pt minus 2pt}

\begin{table}[t]
\Large
\centering%
\renewcommand{\arraystretch}{1.5}
\color{black}
\captionsetup{font={color=black}}
\caption{Time Complexity Comparison of Heterophilous GNNs with Aggregations. $L$ denotes the layer amount, $d$ is the average degree, $k_1 \in \{3,...,7\}$ is the $k_1$ nearest neighbors in U-GCN, $|R|$ is the number of rations in WR-GAT, $k_2 \in\{3,4,5\}$ and $l_{norm}\in\{2,3\}$ are parameters of GloGNN to discover the $k_2$-hop structures and normalization layers in aggregation. }
\vspace{-0.5ex}
\begin{adjustbox}{width=0.5\textwidth}
\begin{tabular}{l|l|l}
\toprule
\textbf{Model} & \textbf{Aggregation} & \textbf{Inference} \\ 
\midrule
{Geom-GCN~\cite{geomgcn}} &$\mathcal{O}(n^2f+mf)$ & $\mathcal{O}(Ln^2f+Lmf+nf^2)$ \\  {GPNN}~\cite{pointer} & $\mathcal{O}(n^2f^2+nf)$ & $\mathcal{O}(n^2f^2+Lmf+nf^2)$ \\  {U-GCN~\cite{cosine}} & $\mathcal{O}(dmf+n^2f+k_1nf)$ & $\mathcal{O}(dmf+n^2f+k_1nf+nf^2)$ \\  {WR-GAT~\cite{structure}} & $\mathcal{O}(Lmf+L|R|n^2f+nf^2)$ & $\mathcal{O}(L|R|n^2f+mf+Lnf^2)$ \\  {GloGNN~\cite{glognn}} & $\mathcal{O}(k_2mfl_{norm})$ & $\mathcal{O}(Lk_2mfl_{norm}+mf+Lnf^2)$ \\  {\textbf{\aggname{}}}(ours) & $\mathcal{O}(knf)$ & $\mathcal{O}(knf+mf+nf^2)$  \\ \bottomrule
\end{tabular}
\label{complexity_comparison}
\end{adjustbox}
\end{table}

It can be interpreted that Algorithm~\ref{simpush} significantly reduces the time complexity for SimRank matrix computation to $\mathcal{O}(d^2)$, which is more scalable with respect to the graph size. To further alleviate time and memory overhead, we utilize top-$k$ pruning to select and store the $k$-largest SimRank scores for each node. Such top-$k$ scheme removes the need of all-pair computation and reduce the space complexity to $\mathcal{O}(kn)$, while maintaining most useful node pair information under the approximation precision~\cite{top-k-simrank}. By addressing the most time-consuming aggregation part of \aggname{}, we then present a thorough analysis highlighting the superiority of \aggname{} complexity among heterophilous GNN aggregations, which is summarized as Table~\ref{complexity_comparison}. Note that models without explicit aggregations such as LINKX are not included. Given a target node $v \in V$ in the $\ell$-th layer, each method conducts: 
\begin{itemize}[wide,labelwidth=!,labelindent=0pt,itemsep=0pt,topsep=0pt]
\item Geom-GCN~\cite{geomgcn} first constructs an extra node set $N_s(v)=\{u \in V|d(h^{(\ell)}_u, h^{(\ell)}_v)\leq \rho\}$, which calculates distance $d(u,v)$ for each node $u\in V$ and retain nodes with their embeddings' distance less than a pre-defined parameter $\rho$, resulting in a $\mathcal{O}(n^2f)$ computation overhead. Combined with original neighbor set $N_g(v)=\{u \in V | (u,v) \in E\}$, it then performs a bi-level aggregation process which takes $\mathcal{O}(n^2f+mf)$ cost in a single aggregation step and $\mathcal{O}(Ln^2f+Lmf)$ with $L$ layers.
% {Geom-GCN calculates all-pair node geometric distance as well as raw edges connectivity, resulting a total complexity of $\mathcal{O}(n^2f+mf)$.} 
\item GPNN~\cite{pointer} initiates the process by using a GCN encoder to establish initial node embeddings, requiring $\mathcal{O}(Lmf)$ time. It subsequently examines all $k$-hop neighbors of a node $v$ for $\forall k\geq 0$, resulting in a node set $N_v$ with a maximum size of $\mathcal{O}(n)$. The method then employs two LSTMs to shuffle the node order before aggregating the features, incurring a total computational cost of $\mathcal{O}(n^2f^2 + nf + mf)$ for the aggregation.
\item U-GCN~\cite{cosine} performs feature aggregation from both its $2$-hop neighbors and its $k_1$-nearest neighbors. This leads to a total time complexity of $\mathcal{O}(dmf+n^2f+k_1nf)$, where $d$ is the average degree, and $k_1$ indicates the $k_1$-nearest neighbors.
\item WR-GAT~\cite{structure} starts by constructing a computational graph $\mathcal{C}$, featuring $\mathcal{O}(m+T\cdot n^2)$ edges, derived from the original graph. Here, $T$ signifies the types of edges included in the graph. Aggregation based on $\mathcal{C}$ then necessitates a computational cost of $\mathcal{O}(mf+T\cdot n^2 f)$ for a single round of aggregation.
{\color{black}\item GloGNN~\cite{glognn} takes a unique approach by employing a well-defined optimization problem to perform feature aggregation. It relies on the $k_2$-hop neighbor structures aand $\ell_{norm}$ normalization layers for this task. Consequently, it demands a time complexity of $\mathcal{O}(k_2mfl_{norm})$ to compute the dominant term $\textbf{A}^{k_2}\cdot \textbf{H}^{(\ell)}$ during each iteration.}
\end{itemize}

{\color{black} For our \aggname{} aggregation following Eq.~(\ref{eq:zsim}), the time complexity is only $\mathcal{O}(knf)$ under the top-$k$ scheme, where $k$ is choice from $\{16,32\}$ to achieve great scalability on large-scale graphs. When the graph grows large, the \aggname{} computation linear to node size $n$, enjoys a better scalability than previous counterparts, as their aggregations with $\mathcal{O}(m)$ or higher complexities quickly become the computational bottleneck~\cite{chiang2019cluster,gbp}. For example, to detect $k_1$ nearest neighbors, U-GCN needs to compute cosine similarities of all node pairs, resulting in $\mathcal{O}(n^2)$ overhead. And for GloGNN, though $k_2$ and $l_{norm}$ are relatively in small range, the aggregation term $\mathcal{O}(k_2l_{norm}mf)$ results in a constant close to $k$, but linear to the edge size $m$, which is considered less scalable of SIGMA. Therefore, we highlight \aggname{} for its superior computation complexity with above analysis.} 
%Consequently, we have the following proposition stating the different part of complexity in \aggname{} as follow:
%\begin{proposition}
%\label{lt}
%    The aggregation time complexity of \aggname{} is $\mathcal{O}(knf)$ and its inference complexity is $\mathcal{O}(knf+mf+nf^2).$
%\end{proposition}

{\color{black}
\subsection{Discussions of \aggname{} with Related Works}
\label{sec:disppr}
In this sub-section, we discuss the relationship and superiority of \aggname{} over existing methods that utilize similarity measurement, such as PPR. We here provide a comprehensive discussion of SIGMA, including similarities and differences with other studies addressing heterophily utilizing global relationships or related matrices.

\mypara{Global Heterophilious GNNs.} SIGMA features the global similarity metric SimRank, which offers distinct interpretation and efficient computation. We list the layer-wise aggregation scheme, final model representation output, and suitability to heterophily for related models in Table~\ref{table:comparison}. The four methods, including H$_2$-GCN \cite{h2gcn}, GPR-GNN \cite{gprgnn}, APPNP \cite{appnp}, and GDC-HK \cite{gasteiger2019}, employ \textit{iterative and hop-by-hop} aggregation in each model layer. H$_2$-GCN integrates features from 1-hop and 2-hop nodes, while the rest gather embeddings from 1-hop neighbors. These models only receive information up to $L$ hops, and in heterophilous scenarios, where global information is useful, these approaches may require many iterations. In contrast, \aggname{} establishes long-term relationships through a single aggregation. The aggregation is more akin to PPRGo \cite{pprgo}, which uses a precomputed relationship matrix based on graph topology. This matrix directly contains the aggregation weights for any node pair $(u,v)$, eliminating the need for iterative propagation. As shown in Fig.\ref{ffig:pagerank} and (c), the SimRank in \aggname{} captures globally similar nodes, whereas PPR matrices perform badly in heterophily contexts. Thus, \aggname{} offers a novel solution to heterophily with global similarity and efficient computation.

\mypara{Walk-Based and Diffusion Approaches.} SimRank can be seen as the meeting probability of pairwise random walks, which is beneficial for capturing homophily between distant nodes, while leveraging efficient one-time precomputation. We highlight its superiority over walk-based methods like PPRGo and diffusion metrics such as GGC-HK in two aspects. First, regarding \textit{global similarity discovery}, the PPR matrix in PPRGo retrieves local information and fails to address heterophily globally. AGP-HK and SGC also act as locality-based band-pass filters, as shown in Fig.\ref{ffig:simrank_diag} from \cite{gasteiger2019}. In contrast, SimRank/SIGMA retrieves global similarity in heterophilous graphs. Second, in terms of \textit{one-time precomputation}, there are no existing methods designed for heterophily graph learning that use a one-time aggregation mechanism. Homophily methods like PPRGo and AGP-HK also underperform in heterophily scenarios. %Our additional experiments comparing these methods with \aggname{} on several heterophily graphs (Table~\ref{table:comparison_pprgo}) show that, while PPRGo is faster in some cases, it consistently achieves suboptimal accuracy. \aggname{} consistently delivers the best performance with comparable or better efficiency, highlighting the limitations of walk-based methods for heterophily.

\mypara{Graph Transformer.} SIGMA's core concept of using global SimRank weights to detect similar nodes and aggregating their representations by a weighted sum aligns with the idea of global attention weights in Graph Transformer (GT). Compared to GT, \aggname{} is powerful for heterophily because of the powerful inductive bias. While GT, whose attention weights are \textit{learnable}, is conceptually more expressive, it needs to be trained without utilizing prior graph knowledge. In fact, this is not so powerful as expected in practice handling graph tasks~\cite{ying2021transformers} and graph information needs to be explicitly incorporated as positional encoding in addition to strengthen its attention weights learning, lacking of clear interpretation and theoretical guarantee towards the practical performance. While for \aggname{}, we've provided religious theoretically guarantee for its performance in handling heterophily graphs. Besides when computing the weights, \aggname{} is linear complexity ($\mathcal{O}(kn)$) compared to the quadratic of GT ($\mathcal{O}(n^2f)$) with the graph node size, which is more applicable to large-scale graphs. This further demonstrates \aggname{}'s superior scalability in handling heterophily graphs.}
%However, in contrast to \aggname{}, Graph Transformers learn attention weights from node features, making them more general and "expressive" for various tasks. While this expressiveness is an advantage, learning attention weights can be more challenging for node classification. These weights are learned from scratch using only node labels as supervision, so there's no guarantee that they will capture complex graph information like SIGMA’s, which is beneficial for heterophily. Moreover, computing learnable attention for all nodes in Graph Transformers is computationally expensive, requiring $\mathcal{O}(n^2f)$ overhead per training epoch, compared to SIGMA's one-time precomputation of weights at $\mathcal{O}(kn)$. This strengthens \aggname{}'s scalability and efficiency.

\begin{table}[t]
\LARGE
\color{black}
\captionsetup{font={color=black}}
\renewcommand{\arraystretch}{1.8}
    \centering
    \caption{Comparison with other global GNNs from \aggname{}.}
    \label{table:comparison}
    \vspace{-0.5ex}
\resizebox{0.5\textwidth}{!}{
    \begin{tabular}{l|l|l}
    \toprule
        \textbf{Model} & \textbf{Layer Aggregation} & \textbf{Final Representation} \\ 
    \midrule
        H$_2$-GCN & $\textbf{H}^{(\ell)}=\textbf{AH}^{(\ell-1)}+\textbf{A}^2\textbf{H}^{(\ell-1)}$ & $\textbf{Z}=\texttt{concat}(\textbf{H}^{(1)},\textbf{H}^{(2)},...,\textbf{H}^{(L)})$ \\ \hline
        GPR-GNN & $\textbf{H}^{(\ell)}= \hat{\textbf{A}}\textbf{H}^{( \ell-1)}$ & $\textbf{Z}= \sum_{ \ell=0}^{L} \gamma_ \ell \textbf{H}^{( \ell)}$ \\ \hline
        APPNP & $\textbf{H}^{( \ell)}=(1- \alpha) \hat{\textbf{A}}\textbf{H}^{( \ell-1)}+ \alpha \textbf{H}^{(0)}$ & $\textbf{Z}=(1- \alpha) \hat{\textbf{A}}\textbf{H}^{(L-1)}+ \alpha \textbf{H}^{(0)}$ \\ \hline
        GDC-HK & $\textbf{H}^{( \ell)}=e^{-t} \frac{t^\ell}{ \ell!}  \hat{\textbf{A}}^ \ell \textbf{H}^{(0)}$ & $\textbf{Z}= \sum_{ \ell=0}^{L}\textbf{H}^{( \ell)}$ \\ \hline
        PPRGo & $\textbf{Z}= \mathbf{\Pi}^{ppr}\textbf{H}^{(0)}$ & $\textbf{Z}= \mathbf{\Pi}^{ppr}\textbf{H}^{(0)}$ \\ \hline
        SIGMA & $\textbf{Z}=\textbf{S} \hat{\textbf{H}}^{(0)}$ & $\textbf{Z}=\textbf{S} \hat{\textbf{H}}^{(0)}$ \\
    \bottomrule
    \end{tabular}}
\end{table}

{\color{black}
\section{Other Related Work}
\label{sec:related_work}
GNNs have made significant advances, largely driven by graph manipulation techniques. Fundamental methods such as GCN~\cite{gnn}, GAT~\cite{gat}, and GraphSAGE~\cite{InductiveRL} utilize graph convolution-based operations. GCNII~\cite{gcn2} deepens the model by stacking network layers for further aggregation, while APPNP~\cite{appnp} and SGC~\cite{wu19sim} decouple network updates into local aggregation and simple feature transformations. Recent works address graph heterophily and the limitations of conventional GNNs, which require global information and distinguishable aggregation schemes. For instance, WRGAT~\cite{structure} introduces new node connections, and GBKGNN~\cite{gbk} and HogGCN~\cite{hog} adjust uniform aggregation by learning homophilic information for identifiable aggregation weights. Additionally, models focused on simple and effective designs for heterophilous graphs have emerged. LINKX~\cite{large}, for example, uses local features and topology with simple \texttt{MLP} layers for embedding construction. Subsequent works~\cite{glognn, zou2023similarity, liang2023predicting} build on \texttt{MLP}(A) to enhance embeddings with additional techniques. GloGNN~\cite{glognn}, while employing complex aggregation schemes requiring multiple iterative rounds to update embeddings, introduces extra computation overhead that makes optimization challenging. In contrast, we highlight \aggname{} for its well-defined topological similarity, efficient computational complexity, and strong theoretical foundation, offering superior performance. As both simple and effective methods of our \aggname{} and LINKX, we distinguish \aggname{} from LINKX that SIGMA introduces a dedicated global aggregation step that leverages the SimRank matrix $\textbf{S}$, capturing distant but similar nodes beyond adjacency, which enhances node homophily representations to above works. This aggregation significantly boosts SIGMA's representational capacity and distinguishes it from LINKX’s purely \texttt{MLP}-based design, with limited performance in tasks requiring message passing. Other works, without \texttt{MLP}(A) component, focus on pure efficiency improvement with PPR calculation~\cite{liao2022scara, liao2023ld2, liao2023scalable, liu2024bird} and dynamic scenarios~\cite{mo2021agenda, zhu2024topology}.}

\section{Experiments}
\label{sec:experiments}
We comprehensively evaluate the performance of \aggname{}, including classification accuracy, efficiency and component study. We also provide a specific study on embedding homophily for aggregating similar nodes globally. The implementation and dataset configurations are available in the GitHub link\footnote{\url{https://github.com/ConferencesCode/SIGMA}}. {\color{black}We conduct all our experiments on a Linux machine with an Intel(R) Xeon(R) Gold 5218 CPU @ 2.30GHz CPU, 256GB RAM and a single Tesla V100 GPU and 40GB memory. All GNN-related training and inference are implemented under PyTorch and the SimRank computation is implemented in C++ and complied by g++ 9.4.0 with -O3 optimization.}

\begin{table*}[!t]\Large
\centering
\caption{The classification accuracy (\%) of \aggname{} and baselines on all datasets. We mark models with average scores ranked \rkat{first}, \rkbt{second}, and \rkct{third} in each dataset. \textbf{Rank} represents the average rankings among all the methods. OOM refers to the out-of-memory error. We also list dataset statistics including homophily value $\mathcal{H}_{node}$ defined in Eq.~(\ref{eq:h_node}).}
\vspace{-0.5ex}
\label{smalltable}
\resizebox{1.0\textwidth}{!}{
\centering
\setlength{\tabcolsep}{1pt}
\renewcommand{\arraystretch}{2.3}
\begin{tabular}{l|cccccccccccc|c}
\toprule[1pt]
\huge \textbf{Dataset} & \huge \ds{Texas} & \huge \ds{Citeseer} & \huge \ds{Cora} & \huge \ds{Chameleon} & \huge \ds{Pubmed} & \huge \ds{Squirrel} & \huge \ds{Genius} & \huge \ds{ArXiv} & \huge \ds{Penn94} & \huge \ds{Twitch} & \huge \ds{Snap} & \huge \ds{Pokec} & \multirow{5}{*}{\huge \ds{Rank}} \\ 
\textbf{\huge Class} & \huge 5 &\huge 6 &\huge 7 &\huge 5 &\huge 3 &\huge 5 &\huge 2 &\huge 5 &\huge 2 &\huge 2 &\huge 5 &\huge 2 &\huge ~ \\
\textbf{\huge Node} &\huge 183 &\huge 3,327 &\huge 2,708 &\huge 2,277 &\huge 19,717 &\huge 5,201 &\huge 421,961 &\huge 169,343 &\huge 41,554 &\huge 168,114 &\huge 2,923,922 &\huge 1,632,803 &\huge ~ \\ 
\textbf{\huge Edge} &\huge 295 &\huge 4,676 &\huge 5,278 &\huge 31,421 &\huge 44,327 &\huge 198,493 &\huge 984,979 &\huge 1,166,243 &\huge 1,362,229 &\huge 6,797,557 &\huge 13,975,788 &\huge 30,622,564 &\huge ~ \\
\textbf{\huge Feature} &\huge 1,703 &\huge 3,703 &\huge 1,433 &\huge 2,325 &\huge 500 &\huge 2,089 &\huge 12 &\huge 128 &\huge 5 &\huge 7 &\huge 269 &\huge 65 &\huge ~ \\ 
\boldmath{\huge $\mathcal{H}_{node}$} &\huge 0.11 &\huge 0.74 &\huge 0.81 &\huge 0.23 &\huge 0.80 &\huge 0.22 &\huge 0.61 &\huge 0.22 &\huge 0.47 &\huge 0.54 &\huge 0.07 &\huge 0.44 &\huge ~ \\ 
\midrule
\huge \texttt{MLP} &{\huge 80.81}±4.7 &{\huge 74.02}±1.9 &{\huge 75.69}±2.0 &{\huge 46.21}±2.9 &{\huge 87.16}±0.3 &{\huge 28.77}±1.5 &{\huge 86.68}±0.1 &{\huge 36.70}±0.2 &{\huge 73.61}±0.4 &{\huge 60.92}±0.1 &{\huge 31.34}±0.1 &{\huge 62.37}±0.1 &{\huge10.3}\\

\huge GAT &{\huge 52.16}±6.6 &{\huge 76.55}±1.2 &{\huge 87.30}±1.1 &{\huge 60.26}±2.5 &{\huge 86.33}±0.5 &{\huge 40.72}±1.5 &{\huge 55.80}±0.8 &{\huge 46.05}±0.5 &{\huge 81.53}±0.5 &{\huge 59.89}±4.1 &{\huge 45.37}±0.4 &{\huge 71.77}±6.1 &{\huge8.9}\\

\huge GBKGNN &{\huge 81.08}±4.8 &{\huge \rkat{79.18}}±0.9 &{\huge 87.29}±0.4 &{\huge 61.59}±2.3 &{\huge 89.11}±0.2 &{\huge 55.90}±1.1 &OOM &OOM &OOM &OOM &OOM &OOM & {\huge 8.8 }\\

\huge HogGCN  & \rkbt{\huge 85.17}±4.4&{\huge 76.15}±1.7 &{\huge 87.04}±1.1 &{\huge 67.27}±1.6 &{\huge 88.79}±0.4 &\rkct{\huge 58.26}±1.5 &OOM &OOM &OOM &OOM &OOM &OOM &{\huge8.7}\\

\huge WRGAT &{\huge 83.62}±5.5 &{\huge 76.81}±1.8 &{\huge 88.20}±2.2 &{\huge 65.24}±0.8 &{\huge 88.52}±0.9 &{\huge 48.85}±0.7 &OOM &OOM &{\huge 74.32}±0.5 &OOM &OOM &OOM &{\huge8.5}\\

\huge H$_2$GCN &{\rkct{\huge 84.16}±7.0} &{\huge 77.11}±1.5 &{\huge 87.87}±1.2 &{\huge 60.11}±2.1 &\rkbt{\huge 89.49}±0.4 &{\huge 36.48}±1.8 &OOM &{\huge 49.09}±0.1 &{\huge 81.31}±0.6 &OOM &OOM &OOM &{\huge 8.2}\\

\huge GPRGNN &{\huge 78.38}±4.3 &{\huge 77.13}±1.6 &{\huge 87.95}±1.2 &{\huge 46.58}±1.7 &{\huge 87.54}±0.4 &{\huge 31.61}±1.2 &{\huge 90.05}±0.3 &{\huge 45.07}±0.2 &{\huge 81.38}±0.2 &{\huge 61.89}±0.3 &{\huge 40.19}±0.1 &{\huge 78.83}±0.1 &{\huge 8.0}\\

\huge GCN &{\huge 55.14}±5.1 &{\huge 76.50}±1.3 &{\huge 86.98}±1.2 &{\huge 64.82}±2.2 &{\huge 88.42}±0.5 &{\huge 53.43}±2.0 &{\huge 87.42}±0.3 &{\huge 46.02}±0.2 &{\huge 82.47}±0.2 &{\huge 62.18}±0.2 &{\huge 45.65}±0.0 &{\huge 75.45}±0.1 &{\huge 7.6 }\\

\huge ACMGCN &{\huge 84.67}±4.3 &{\huge 77.13}±1.7 &{{\huge 87.91}}±0.9&{{\huge 66.93}}±1.8 &{\huge 89.17}±0.52 &{{\huge 54.40}±1.8}&{\huge 80.33}±3.9 &{\huge 47.16}±0.6 &{\huge 82.52}±0.9 &{\huge 62.01}±0.7 &{\huge 55.14}±0.1 &{\huge 63.81}±5.2&{\huge 6.9}\\

\huge MixHop &{\huge 77.84}±7.7 &{\huge 76.26}±1.3 &{\huge 87.61}±0.8 &{\huge 60.50}±2.5 &{\huge 85.31}±0.6 &{\huge 43.80}±1.4 &{\huge 90.58}±0.1 &{\huge 51.81}±0.1 &{\huge 83.47}±0.7 &{\huge 65.64}±0.2 &{\huge 52.16}±0.1 &{\huge 81.07}±0.1 &{\huge 6.6}\\

\huge GCNII &{\huge 77.57}±3.8 &\rkct{\huge 77.33}±1.4 &{\rkbt{\huge 88.37}}±1.2 &{\huge 63.86}±3.0 &{\rkct{\huge 89.36}±0.3} &{\huge 38.47}±1.5 &{\huge 90.24}±0.1 &{\huge 47.21}±0.2 &{\huge 82.92}±0.5 &{\huge 63.39}±0.6 &{\huge 47.59}±0.6 &{\huge 78.94}±0.1 &{\huge 5.7}\\

\huge LINKX &{\huge 74.60}±8.3 &{\huge 73.19}±0.9 &{\huge 84.64}±1.1 &\rkct{\huge 68.42}±1.3 &{\huge 87.86}± 0.7 &{\rkbt{\huge 61.81}}±1.8 &\rkct{\huge 90.77}±0.2 &{\huge \rkat{56.00}}±1.3 &\rkct{\huge 84.71}±0.5 &\rkct{\huge 66.06}±0.2 &\rkct{\huge 61.95}±0.1 &\rkct{\huge 82.04}±0.1 &{\huge 5.5}\\

\huge GloGNN &{\huge 84.05}±4.9 &{\huge 77.22}±1.7 &\rkct{\huge 88.33}±1.0 &{\rkbt{\huge 71.21}±1.8} &{\huge 89.24}±0.4 &{\huge 57.88}±1.7 &{\rkbt{\huge 90.91}±0.1} &\rkct{\huge 54.79}±0.2 &{\rkbt{\huge 85.74}±0.4} &{\rkbt{\huge 66.34}±0.3} &{\rkbt{\huge 62.03}±0.2} &{\huge \rkat{83.05}}±0.1 &{\rkbt{\huge 2.9} }\\ 

\huge \textbf{\aggname{}} &{\huge \rkat{85.32}}±4.7 &{\rkbt{\huge 77.52}±1.5} &{\huge \rkat{88.96}}±1.2 &{\huge \rkat{72.13}}±1.7 &{\huge \rkat{89.76}}±0.3 &{\huge \rkat{62.04}}±1.6 &{\huge \rkat{91.68}}±0.6 &{\rkbt{\huge 55.16}±0.3} &{\huge \rkat{86.31}}±0.3 &{\huge \rkat{67.21}}±0.3 &{\huge \rkat{64.63}}±0.2 &{\rkbt{\huge 82.33}}±0.1 &{\huge \rkat{1.2} }\\ \bottomrule[1pt]
\end{tabular}}

\vspace{-0.5ex}
\end{table*}

\subsection{Experiment Setup}
\mypara{Baselines.}
We compare \aggname{} with 12 baselines. (1) General graph models: MLP, GCN~\cite{gnn}, GAT~\cite{gat}, Mixhop~\cite{mixhop}, and GCNII~\cite{gcn2}; (2) Graph convolution-based heterophilous models: H$_2$GCN~\cite{h2gcn}, GPRGNN~\cite{gprgnn}, WRGAT~\cite{structure}, ACMGCN~\cite{acmgcn}, HogGCN~\cite{hog} and GBKGNN~\cite{gbk}; (3) Decoupled heterophilous models: LINKX~\cite{large} and GloGNN~\cite{glognn}.

\begin{table}[!t]
\centering
\caption{Parameter search ranges.}
\vspace{-0.5ex}
\label{hyper_small}
\renewcommand{\arraystretch}{1.6}
\resizebox{0.47\textwidth}{!}{\begin{tabular}{l|c|c}
\toprule
\textbf{Parameter}&  \textbf{small-scale\cite{geomgcn}} & \textbf{large-scale\cite{nonlocal}}  \\ 
\midrule
decay factor $c$ &[0,1] &[0,1] \\
feature factor $\delta$ & \{0, 0.05, 0.1, ..., 1\} & \{0, 0.5, 1\} \\ 
dropout $p$ & \{0, 0.05, 0.1, ..., 0.95\} & \{0, 0.05, 0.1, ..., 0.95\} \\ 
learning rate $r$ & [1e-6, 1e-2, 1e-4] & \{1e-2, 1e-3,1e-4, 1e-5\} \\ 
weight decay & [0, 1e-2, 1e-3, 1e-4] & \{0, 1e-3, 1e-4, 1e-5\} \\ 
hidden layer & \{32, 64, 128, 256\} & \{32, 64, 128, 256, 512\} \\ 
early stopping & \{40, 100, 150, 200, 300\} & \{100, 300\} \\ 
\bottomrule
\end{tabular}}
\vspace{-0.5ex}
\end{table}

\mypara{Datasets} 
We conduct experiments on 12 real-world datasets, including 6 small-scale and 6 large-scale datasets spanning various domains, features, and graph homophiles. Among them, \textit{Texas}, \textit{Chameleon}, and \textit{Squirrel}~\cite{dataset1} are three webpage datasets collected from Wikipedia or Cornell University with low homophily ratios. \textit{Cora}, \textit{Citeseer}, \textit{Pubmed}, \textit{Arxiv-year} and \textit{Snap-patents}~\cite{cora, citeseer, pubmed, arxiv-year, snap-patents} are citation graphs where the first three have high and the last two have low homophily ratios. \textit{Penn94}, \textit{pokec}, \textit{Genius}, and \textit{twitch-gamers}~\cite{large} are all social networks extracted from online websites such as Facebook, having moderate homophily ratios. The statistics containing number of nodes, edges, categories, and features, as well as homophily ratios are summarized in Table \ref{smalltable}.

\mypara{Parameter Settings.}
{\color{black} We calculate exact SimRank score for small datasets, while adopting approximation as described in Section~\ref{ssec:scale} with $\epsilon = 0.1$ and $k\in\{16,32\}$ on large-scale graphs}, which is sufficient to derive high-quality aggregation coefficients (see Section~\ref{section:ablation_study} for detail).
The initial coefficient $\alpha = 0.5$ on all the datasets. The layer number of \textit{\texttt{MLP}}$_{H}$ is set to 1 and 2 for small and large datasets. We use the same dataset train/validation/test splits as in~\citet{glognn}. We conduct 5 and 10 repetitive experiments on the small and large datasets, respectively. Detailed exploration on parameters including feature factor $\delta$, learning rate $r$, dropout $p$ and weight decays are elaborated in Table~\ref{hyper_small}.

\subsection{Performance Comparison}
\label{section:performance}

We measure our \aggname{}'s performance against 12 baselines on 12 benchmark datasets of both small and large scales in Table~\ref{smalltable}, where we also rank and order the models based on respective accuracy. We stress the effectiveness of \aggname{} by the following observations:

\mypara{General GNNs.} General graph learning models perform worse in most cases. Among MLP, GAT, and GCN, GCN achieves the highest ranking 7.67 over all datasets. The reason may be that they fail to distinguish the homophily and heterophily nodes due to the uniform aggregation design. It is surprising that MLP only utilizing node features learns well on some datasets such as \emph{Texas}, which indicates that node features are important in heterophilous graphs' learning. Meanwhile, Mixhop and GCNII generally hold better performance than the plain models. On \emph{pokec}, the accuracy for Mixhop and GCNII are 81.07 and 78.94, outperforming the former three models significantly. They benefit from strategies considering more nodes in homophily and combine node representations to boost the performance, showing the importance of modifying the neighborhood aggregation process. 

\begin{table*}[!t]
\large
\centering
\color{black}
\captionsetup{font={color=black}}
\setlength{\tabcolsep}{6pt}
\caption{The average learning time (s) on large-scale datasets. We separately show the break down of precomputation (Pre.) and aggregation (AGG) when applicable. The overall learning efficiency are marked with \rkat{first}, \rkbt{second} and \rkct{third} place. The average speed-up of \aggname{} over other methods are shown in Avg.$\uparrow$.}
%\vspace{-0.5ex}
\label{learntime}
\vspace{-0.5ex}
\setlength{\tabcolsep}{3pt}
\renewcommand{\arraystretch}{1.2}
\resizebox{\textwidth}{!}{
\begin{tabular}{l|ccc|ccc|ccc|ccc|ccc|ccc|c}
\toprule
\multirow{2}{*}{\textbf{Model}} & \multicolumn{3}{c|}{\ds{genius}} & \multicolumn{3}{c|}{\ds{arXiv}} & \multicolumn{3}{c|}{\ds{Penn94}} & \multicolumn{3}{c|}{\ds{twitch}} & \multicolumn{3}{c|}{\ds{snap}} & \multicolumn{3}{c|}{\ds{pokec}}& \multirow{2}{*}{Avg.$\uparrow$}\\ 
  ~ & { Pre.} & { AGG} & { Learn} 
    & { Pre.} & { AGG} & { Learn} 
    & { Pre.} & { AGG} & { Learn} 
    & { Pre.} & { AGG} & { Learn} 
    & { Pre.} & { AGG} & { Learn} 
    & { Pre.} & { AGG} & { Learn} & \\
\midrule
  LINKX &- &- & \rkbt{292.3} &- &- & \rkbt{51.2} &- &- & \rkbt{49.6} &- &- & \rkbt{302.7} &- &- & \rkbt{469.1} &- &- &\rkbt{672.3} & $1.73\times$  \\
  GloGNN &- & 313.6& \rkct{358.7}&- & 126.1 & \rkct{134.1} &- & 167.1 & \rkct{183.5} &- & 739.5 & \rkct{783.0} &-&696.3 &\rkct{732.9} &- & 1417.8 & \rkct{1564.7} & $4.30\times$ \\
  \aggname{} &8.6 &119.8 & \rkat{153.6}&9.3 &26.6 & \rkat{36.5} &3.9 &6.9 & \rkat{17.2} &14.0 & 189.2 & \rkat{236.5} &15.9 & 314.3 & \rkat{408.2} &11.4 & 279.7 &\rkat{388.5} & - \\
\bottomrule
\end{tabular}
}
\vspace{-2.ex}
\end{table*}

\mypara{Heterophilous GNNs.} With respect to heterophilous models, those graph convolution-based approaches enjoy proper performance on small graphs by incorporating structural information, but their scalability is a bottleneck. Specifically, H$_{2}$GCN, and WRGAT cannot be employed in most large datasets due to their high memory requirement for simultaneously processing features and conducting propagation, which hinders their application. Meanwhile, both HogGCN and GBKGNN fail to run on all the large-scale datasets as they require much more parameters to learn homophily properties. GPRGNN, on the contrary, shows no superiority in effectiveness. On the other hand, decoupling heterophilous models LINKX and GloGNN achieve the most competitive accuracy over most datasets where LINKX separately embeds node features and graph topology and GloGNN further performs node neighborhood aggregation from the whole set of nodes in the graph, bringing them performance improvement. Notably, the advantage of LINKX is not consistent and usually better on larger graphs. 

\mypara{\aggname{}.} Our proposed method, \aggname{}, consistently outperforms competing algorithms across a diverse range of datasets, achieving the highest average accuracy on nine out of twelve tested datasets. Specifically, it excels in the top ranking score of 1.25, performing a significant advantage over its closest competitor, GloGNN as 2.75. This exceptional performance is attributed to our innovative global aggregation strategy, which effectively leverages structural information to evaluate node similarity. Moreover, our decoupled feature transformation network architecture significantly enhances the extraction of node features and the generation of meaningful embeddings. For instance, on the \textit{snap-patents} dataset, \aggname{} surpasses runner-up methods by a margin exceeding 1.5\%, highlighting its superior ability to differentiate between homophily and heterophily. However, it is worth noting that \aggname{} slightly lags behind the best-performing method in \emph{arXiv-year} and \emph{pokec}. This under-performance is attributed to its shallow feature transformation layers, which may lack the capacity to effectively handle large volumes of input information. %Further analyses, including potential remedies for this limitation, are discussed in detail in Appendix~\ref{section:discussion}.

\subsection{\color{black}Scalability and Efficiency Study}
\label{sec:exp:scalability}

\begin{figure}[t]
\centering
\vspace{-1.0ex}
\includegraphics[width=0.48\textwidth]
{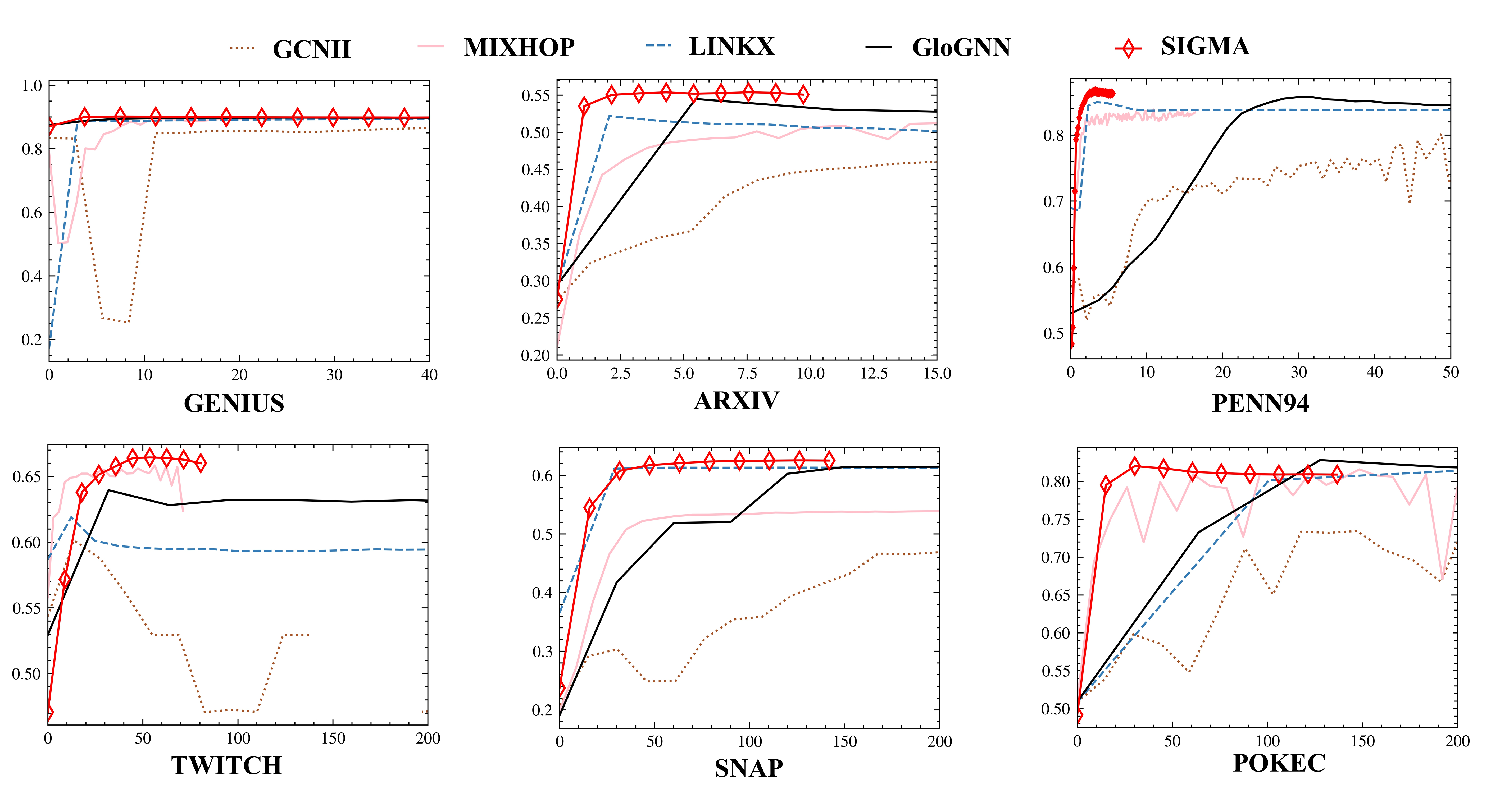}
\vspace{-0.5ex}
\caption{Convergence efficiency of 
\aggname{} and leading baselines. X-axis denotes the training time (s) and Y is the accuracy (\%).}
\vspace{-0.5ex}
\label{efficiency}
\end{figure}

\begin{figure}[t]
\centering
\centering
\vspace{0.6ex}
\includegraphics[width=0.42\textwidth]{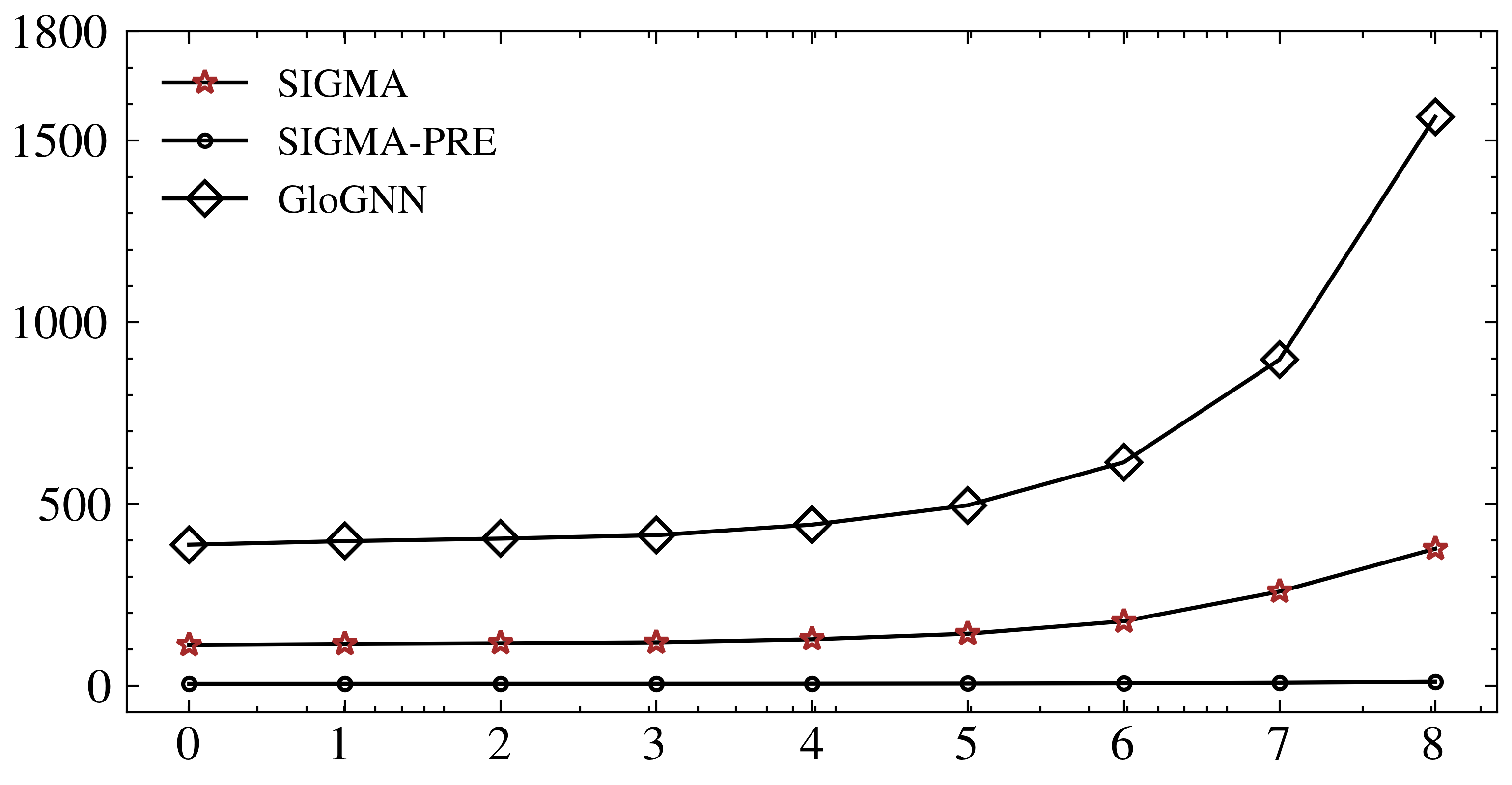}
\captionsetup{font={color=black}}
\vspace{-0.5ex}
\caption{Scalability Evaluation of SIGMA and GloGNN. X-axis denotes the graph edge scale at $\{\frac{3\times 10^8}{2.5^i}\}_{i=0}^8$  and Y-axis the wall clock time (s). Note that X-axis is in \textit{log-scale}.}
\vspace{-1.ex}
\label{fig:scalability}
\end{figure}

This section studies the efficiency of \aggname{}. To evaluate the efficiency aspect of \aggname{}, we investigate its learning time and convergence speed in Table~\ref{learntime} and Fig.\ref{efficiency}. And we validate the scalability of \aggname{} in Fig.\ref{fig:scalability}.
\mypara{Learning Time.} We compare the learning time of SIMGA with LINKX and GloGNN as they are all decoupling heterophilous methods and achieve most competitive performance among baselines. We denote learning time the summation of pre-computation time and training time for a fair comparison since \aggname{} needs to calculate the aggregation matrix before conducting network training. We use the same training set on each dataset and run 5-repeated experiments for 500 epochs. Note that for each method, we utilize configurations and parameters corresponding to their first-tier performance after tuning. The average learning time is reported in Table \ref{learntime}, along with the aggregation part. It can be seen that \aggname{} costs the least learning time over all the large-scale datasets, thanks to its scalable pre-computation and top-$k$ global aggregation mechanism, which aligns with our complexity analysis in Section~\ref{ssec:scale}. Besides, the aggregation costs much less time for \aggname{} due to the one-time similarity measurement calculation design, compared to GloGNN's to-be-updated measurement. On datasets such as \emph{Penn94} and \emph{pokec}, \aggname{} outperforms GloGNN by around $10\times$ and $5\times$ faster learning time correspondingly, which greatly demonstrates the efficiency and scalablity of our aggregation method. Concerning LINKX, its classification performance is outperformed by \aggname{} largely, which severely hinders its wide applications. %We also observe that SIGMA is faster than LINKX, despite incorporating additional global aggregation. This improvement is attributed to a slight but effective modification: while LINKX concatenates the raw node and adjacency features as $\textbf{H} = \texttt{MLP}(\texttt{concat}[\mathbf{H_A}, \mathbf{H_X]})$, SIGMA introduces a tunable parameter $\delta$ to balance the combination as $\textbf{H} = \texttt{MLP}(\delta \cdot \mathbf{H_A} + (1 - \delta) \cdot \mathbf{H_X})$. This design not only allows for flexible adjustments but also reduces the initial hidden size by half, resulting in a $4\times$ improvement in running efficiency within the hidden layer.

{\color{black}
\mypara{Convergence.}
We study the convergence time as an indicator of model efficiency in graph learning. We compare \aggname{} against leading baselines, including MixHop, GCNII, LINKX, and GloGNN, in Fig.\ref{efficiency}. The results show that \aggname{} achieves favorable convergence, attaining high accuracy in a short training time. Generally, \aggname{}, MixHop, and LINKX converge quickly to their highest scores due to their simple designs, with \aggname{} often achieving better ultimate accuracy and maintaining it more consistently than LINKX and MixHop. Additionally, \aggname{} demonstrates clear advantages over GloGNN: it converges faster and reaches higher final accuracy. For instance, its convergence acceleration over GloGNN reaches $10\times$ on the graph \emph{Penn94}. These results validate that \aggname{} is both highly effective and efficient, making it well-suited for large-scale heterophilous graphs with superior scalability.}

{\color{black}
\mypara{Scalability.}
To clearly evaluate the scalability of SIGMA, we here vary the graph scales to observe how the learning and pre-computation time of \aggname{} changes. We also include the best baseline GloGNN for chasing the speed-up trends. Specifically, we use the largest graph, \textit{pokec}, with roughly $3 \times 10^8$ edges as our base-graph. We then vary the graph size by randomly removing or adding edges. To maintain clarity regarding the graph scale, we construct a series of graphs $\{g\}_{i=0}^8$, where the $i$-th graph contains $\frac{3 \times 10^8}{2.5^i}$ edges, respectively. As a result, the graph size scales from roughly $2 \times 10^5$ edges to $3 \times 10^8$ edges. We then conduct additional experiments with the same settings and report the time comparison in Fig.\ref{fig:scalability}. Based on the results, we can see that (1) In general, the construction time of SIGMA remains relatively low and steady compared to the overall learning time and (2) Both SIGMA's and GloGNN's learning times scale linearly with the number of graph edges (note the x-axis is log-scaled). The overall trend shows that SIGMA's speed-up over GloGNN increases as the graph size grows. Therefore, we emphasize SIGMA's superior scalability over GloGNN, especially for larger graphs.}

\begin{figure}[!t]
\centering
\vspace{-1ex}
{\includegraphics[width=0.4\textwidth]{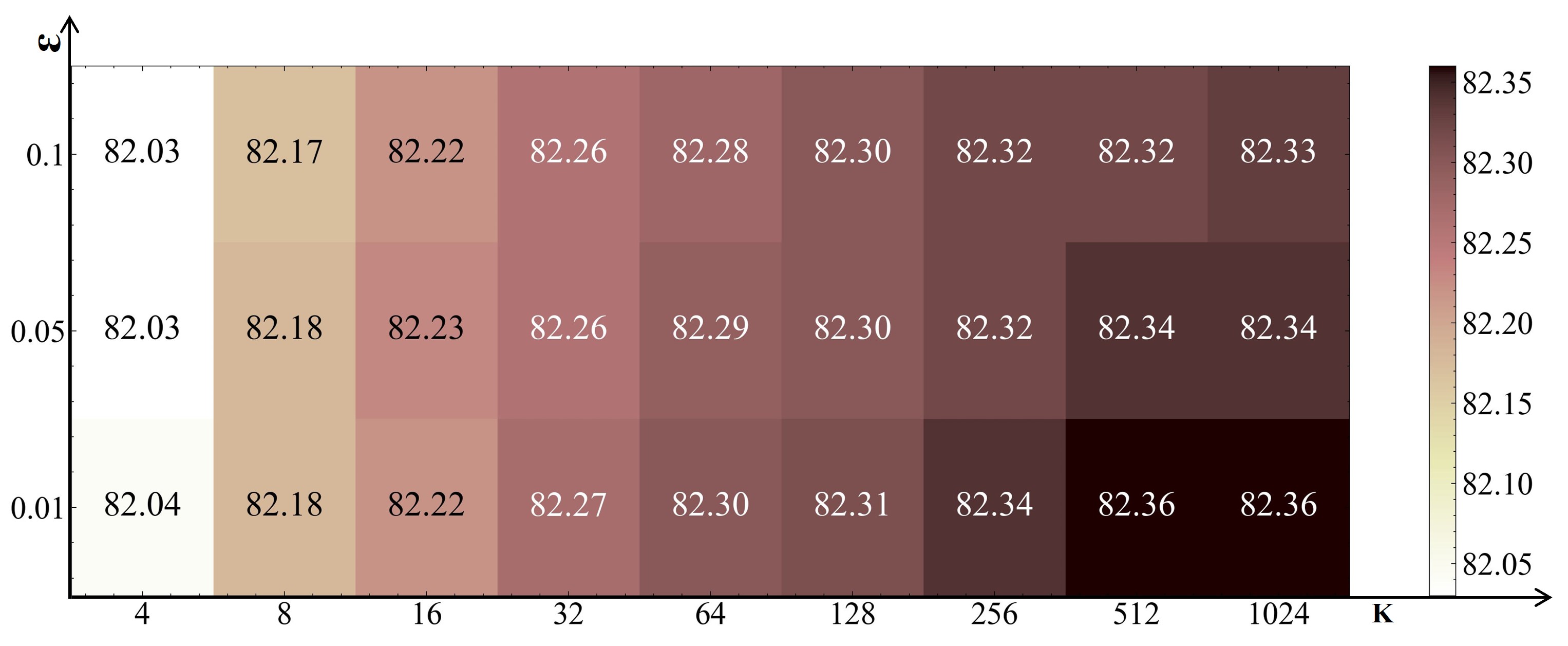}}
\vspace{-0.5ex}
\caption{Effect of error parameter $\epsilon$ and top-$k$ on graph \textit{pokec}.}
%\vspace{-0.5ex}
\label{fig:abalation_b}
\end{figure}

\subsection{Components Evaluation}
\label{section:ablation_study}
In this section, we conduct additional experiments to explore the influence and choice of \aggname{}'s components, approximation factors and several hyper-parameters. 

\begin{table*}[!t]
\color{black}
\tiny
\renewcommand{\arraystretch}{1.2}
\captionsetup{width=\textwidth}
\vspace{-0.5ex}
\caption{%\haoyu{add local SIGMA with $S\otimes A$ aggregation.}
Component study on the effect of \aggname{} and GloGNN components $\mathbf{S}$, $\mathbf{S\cdot A}$, $\mathbf{A}$, and $\mathbf{X}$. Avg.$\downarrow$ and Max.$\downarrow$ denote the average and maximum accuracy drop of the model respectively. }
%\vspace{-0.5ex}
\label{tab:ablation}
\begin{adjustbox}{width=\textwidth, center}
\begin{tabular}{p{1.8cm}|P{0.8cm}P{0.8cm}P{0.8cm}P{0.8cm}P{0.8cm}P{0.8cm}|P{0.8cm}P{0.8cm}}
\toprule
{\textbf{Component}} & \ds{Genius} & \ds{ArXiv} & \ds{Penn94} & \ds{Twitch} & \ds{Snap} & \ds{Pokec} & \ds{Avg.$\downarrow$} & \ds{Max.$\downarrow$} \\ 
\midrule
{\aggname{}} & 91.68 & 55.16 & 86.31 & 67.21 & 64.63 & 82.33 & - & - \\ 
{\aggname{} w/o $\mathbf{S}$} & 87.21 & 53.67 & 83.72 & 63.81 & 61.96 & 81.78 & 2.51 & 3.73 \\ 
{\aggname{} w/ $\mathbf{S\cdot A}$} & 90.37 & 53.89 & 84.59 & 64.56 & 62.73 & 81.27 & 1.65 & 2.65 \\
{\aggname{} w/o $\mathbf{X}$} & 73.84 & 53.10 & 82.46 & 62.83 & 62.34 & 81.92 & 4.78 & 17.2 \\
{\aggname{} w/o $\mathbf{A}$} & 86.08 & 34.97 & 75.36 & 60.01 & 36.26 & 60.16 & 15.4 & 25.7 \\ 
GloGNN & 0.9066& 0.5468& 0.8557& 0.6619& 0.6209& 0.9300 & - & - \\ 
\midrule
{GloGNN w/o $\mathbf{A}$} & 0.7463& 0.3705& 0.7978& 0.5987& 0.3066& 0.6220 & 16.3 & 31.43 \\ 
{GloGNN w/o $\mathbf{X}$} & 0.8718& 0.5445& 0.8203& 0.6559& 0.6153& 0.8203 & 1.5 & 3.53 \\ 
\bottomrule
\end{tabular}
\end{adjustbox}
\vspace{-4ex}
\end{table*}

\mypara{Effect of $\mathbf{S}$, $\mathbf{S\cdot A}$, $\mathbf{A}$, and $\mathbf{X}$ in \aggname{}.} Recalling Equation~(\ref{eq15}), the parameter $\delta$ is used to balance the contributions of the attribute matrix $\mathbf{X}$ and the adjacency matrix $\mathbf{A}$. Setting $\delta=0$ and $\delta=1$ isolates the effects of $\mathbf{A}$ and $\mathbf{X}$, respectively, with the corresponding configurations denoted as \aggname{} w/o $\mathbf{X}$ and \aggname{} w/o $\mathbf{A}$. Additionally, setting $\alpha=1$ in Equation~(\ref{eq8}) allows us to analyze \aggname{} w/o $\mathbf{S}$, examining the effectiveness of the core aggregation component.

To further investigate whether the global properties of $S$ in \aggname{} aid learning, we develop a localized version by substituting the aggregation matrix $\mathbf{S}$ with $\mathbf{S} \cdot \mathbf{A}$, referred to as \aggname{} w/ $\mathbf{S} \cdot \mathbf{A}$. This confines the effective coefficients to immediate neighbors derived from the original adjacency matrix $\mathbf{A}$. We evaluate these configurations across large-scale graphs presented in Table~\ref{tab:ablation} and we conclude:

\begin{enumerate}[wide,labelwidth=!,labelindent=0em,itemsep=1pt,topsep=1pt]
    \item \textbf{Integration of $\mathbf{S}$ improves performance.} Incorporating $\mathbf{S}$ into the aggregation process significantly enhances classification accuracy, with an average improvement margin of 2.51\%. This highlights the efficacy of \aggname{} and underscores the importance of its robust design in boosting performance.

    {\color{black}
    \item \textbf{Both $\mathbf{A}$ and $\mathbf{X}$ are essential components.} The removal of either $\mathbf{A}$ or $\mathbf{X}$ greatly degrades classification performance, particularly for $\mathbf{A}$. And we find that both \aggname{} and GloGNN encounter such conditions. This indicates that the \texttt{MLP}(A) is a fundamental component in initializing representative node embeddings before further processed. As such, structural information present in $\mathbf{A}$ can be further combined with for example $S$ iin \aggname{} to obtain improved performance.}

    \item \textbf{Global aggregation functions are critical.} The accuracy decline observed with \aggname{} w/ $\mathbf{S} \cdot \mathbf{A}$ emphasizes the importance of global aggregation capabilities, such as those facilitated by SimRank. This aligns with our theoretical interpretation in Section~\ref{cor}, which identifies the importance of long-distance node pairs enriched with additional homophily information. Restricting nodes detected through $S$ to immediate neighbors significantly reduces performance.
\end{enumerate}

\begin{figure}[!t]
\centering
\vspace{-0ex}
\includegraphics[width=0.45\textwidth]{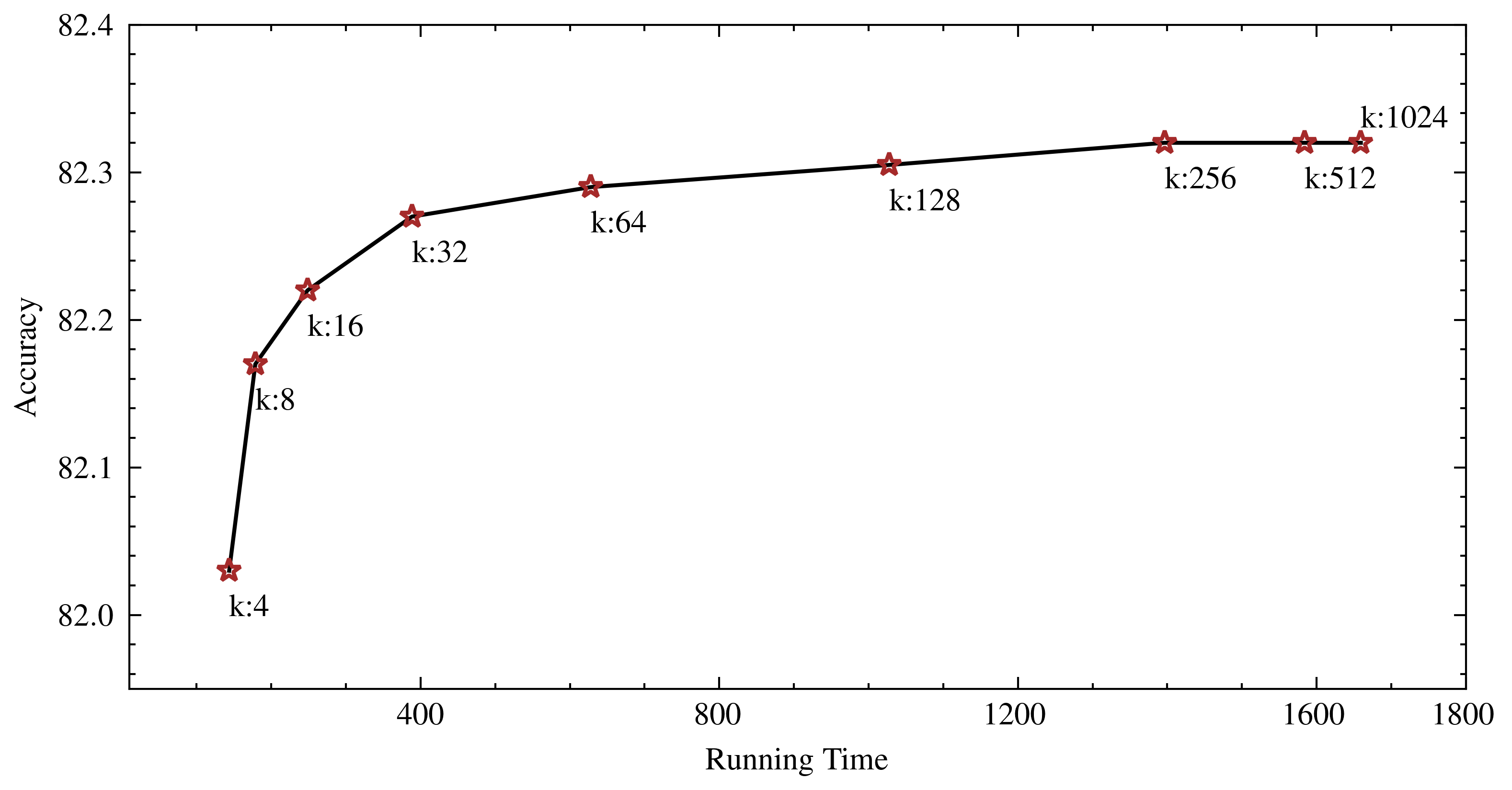}
\vspace{-0.5ex}
\captionsetup{font={color=black}}
\caption{Trade-offs of runtime over the top-$k$ scheme. The X-axis represents the actual runtime, and the Y-axis represents accuracy. The value of $k$ is indicated on each data point.}
\label{fig:top_k_tradeoff}
\vspace{-0.5ex}
\end{figure}

{\color{black}
\mypara{Practical Choice of Parameter $\mathbf{k}$ and $\boldsymbol{\epsilon}$.}  The choice of $\epsilon$ and top-$k$ directly impacts the efficiency of \aggname{}, along with the performance. The parameter $\epsilon$ controls the approximation error of $S$, affecting the pre-computation time, while top-$k$ determines the number of nodes selected for aggregation. In general, smaller values of $\epsilon$ and larger values of $k$ increase computational complexity. On our empirical evaluation in Fig.\ref{fig:abalation_b}, we vary $k \in \{4, 8, \ldots, 1024\}$ and set $\epsilon \in \{0.01, 0.05, 0.1\}$. Our observations reveal that $\epsilon = 0.1$ provides satisfactory classification performance for the largest dataset, \textit{pokec}. More stringent settings, such as $\epsilon = 0.01$, yield only marginal performance improvements while increasing pre-computation time. This demonstrates that a rough approximation with $\epsilon = 0.1$ allows for an efficient pre-computation stage, significantly speeding up \aggname{}. Regarding top-$k$ in Fig.\ref{fig:top_k_tradeoff}, we fix $\epsilon=0.1$ and evaluate the trade-off on accuracy and runtime. We can see that the accuracy increases as $k$ grows, but the improvements become smaller. Initially, there is a noticeable increase in accuracy as $k$ increases, but after surpassing a certain threshold ($k = 32$), the accuracy stabilizes with only minor incremental improvements. The accuracy gains increase by only around 0.05 from $k = 32$ to $k = 1024$, while the runtime nearly triples. This suggests that the decision to increase $k$ should balance the computational cost with the diminishing returns in accuracy. Thus we practically choose $k \in \{16, 32\}$ for large-scale graphs to achieve a good trade-off.}

\mypara{Sensitivity of $\mathbf{\delta}$ and $\mathbf{\alpha}$.} The parameter $\delta$ controls the balance between raw node features and adjacency matrix features. Table~\ref{tab:ablation} shows that extreme values of $\delta=1$ and $\delta=0$ demonstrate the importance of both factors, as both contribute significantly to performance. The sensitivity of $\delta$ varies across datasets, with datasets like \emph{PENN94} showing a preference for a setting slightly biased towards the adjacency matrix, while \emph{POKEC} favors node features. Varying $\delta$ in $[0.1, 0.3, 0.5, 0.7, 0.9]$, as seen in Table~\ref{tab:sense_d}, shows that SIGMA adapts effectively to the varying importance of feature and structure information in different graphs. The balance parameter $\alpha \in [0,1]$ manages the trade-off between global and local aggregation representations. Smaller values of $\alpha$ prioritize global aggregation from $S$. Table~\ref{tab:alpha_values} reports convergent values of $\alpha$ across six datasets, showing that optimal $\alpha$ values vary slightly. Smaller $\alpha$ values are more effective when global aggregation is critical. For the heterophilous \textit{SNAP} graph, the learned $\alpha$ is much smaller, highlighting the importance of SIGMA's global aggregation in handling extreme heterophily.

\begin{table}[!t]
\tiny
\caption{Results for different datasets across $\delta$ values.}
\vspace{-0.5ex}
\centering
\renewcommand{\arraystretch}{1}
\setlength{\tabcolsep}{11pt}
\resizebox{0.48\textwidth}{!}{
\begin{tabular}{c|ccc}
    \toprule
    Value of $\mathbf{\delta}$ & \ds{PENN94} & \ds{ARXIV} & \ds{POKEC} \\
    \midrule
    $0.1$ & 82.91 & 55.16 & 81.81 \\
    $0.3$ & 84.14 & 54.74 & \textbf{82.33} \\
    $0.5$ & 85.20 & 54.78 & 82.13 \\
    $0.7$ & \textbf{86.21} & \textbf{54.91} & 81.97 \\
    $0.9$ & 84.97 & 54.53 & 81.81 \\
    \bottomrule
\end{tabular}
}
%\vspace{-0.5ex}
\label{tab:sense_d}
\end{table}

\begin{table}[!t]
\tiny
\caption{Values of convergent $\alpha$ on six large-scale datasets.}
\vspace{-0.5ex}
\centering
\renewcommand{\arraystretch}{1.2}
\setlength{\tabcolsep}{5pt}
\resizebox{0.48\textwidth}{!}{
\begin{tabular}{ccccccc}
    \toprule
     & \ds{GENIUS} & \ds{ARXIV} & \ds{PENN94} & \ds{TWITCH} & \ds{SNAP} & \ds{POKEC} \\
    \midrule
    $\mathbf{\alpha}$ & 0.52 & 0.28 & 0.33 & 0.34 & 0.12 & 0.48 \\
    \bottomrule
\end{tabular}
}
\label{tab:alpha_values}
\end{table}

\begin{figure}[!b]
  \centering
  \vspace{-0.5ex}
  \includegraphics[width=.48\textwidth]{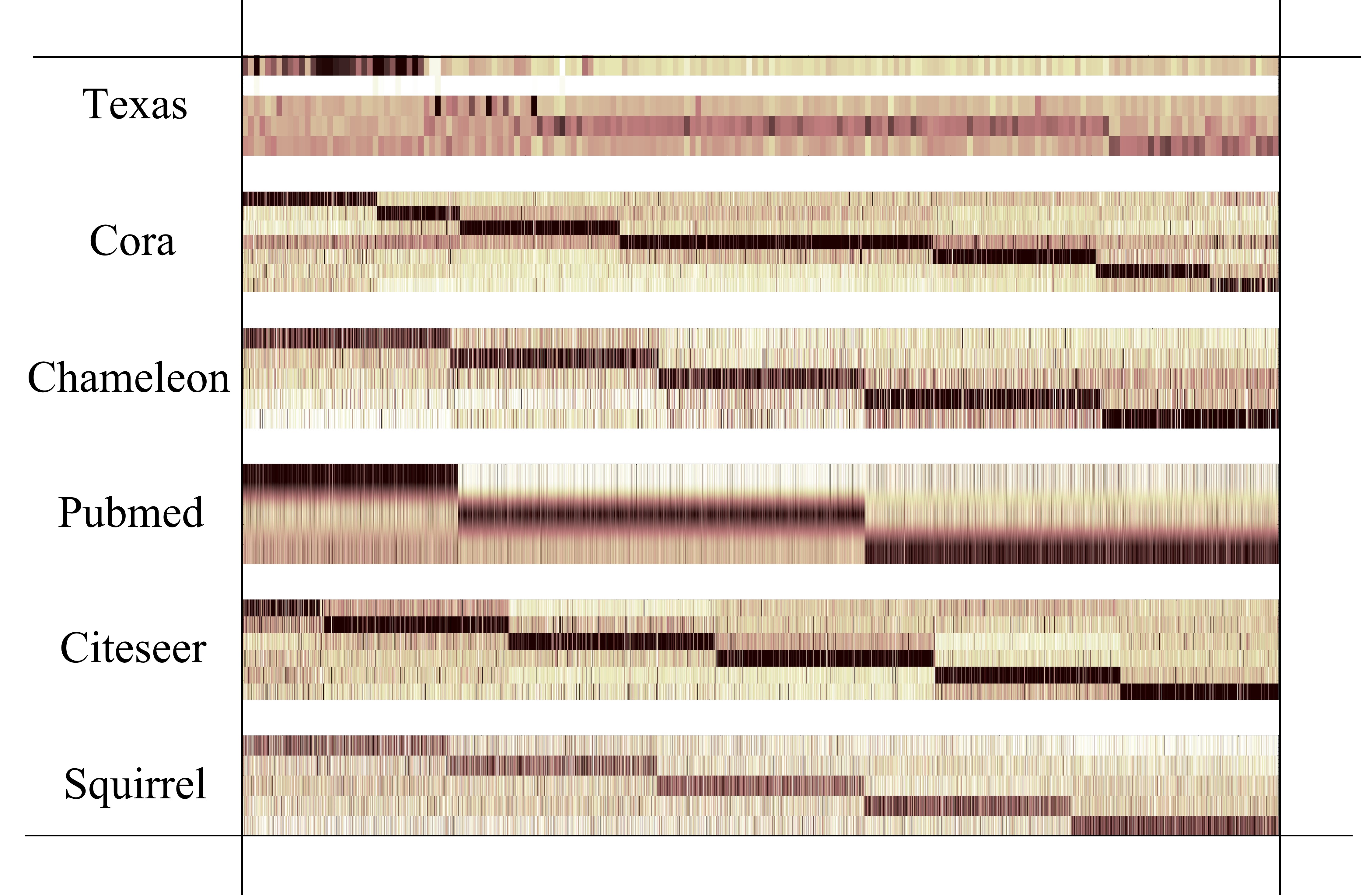}
  \vspace{-0.5ex}
  \caption{Homophily in node embeddings $Z$. X-axis corresponds to node index reordered by category labels, color along Y-axis represents values in the node embedding vector.}
  \label{group}
\vspace{-0.5ex}
\end{figure}

\subsection{Grouping Effect Visualization}
\label{sec::homo}
To demonstrate the capabilities of our model, we present the output embedding matrix $\textbf{Z}$ (Eq.~(\ref{eq8})) and show its application on six small-scale graphs in Fig.\ref{group}. As shown in Theorem~\ref{t2}, the $\textbf{Z}$ matrix exhibits the desired grouping effect, confirmed visually in Fig.\ref{group}. Nodes with the same category labels have similar embedding patterns, while nodes from different categories show clear distinctions. For example, in the \textit{Pubmed} dataset (Fig.\ref{group}), which has three node types, the embeddings form distinct rectangular patterns. The number of patterns corresponds to the number of categories, demonstrating \aggname{}'s ability to capture global homophily and distinguish heterophily. These observations provide strong evidence of \aggname{}'s effectiveness in capturing complex graph structures.

\begin{table}[!t]
\centering
% \Large
\caption{Exploration on iterative \aggname{}.}
\label{tab:iterative}
\vspace{-0.5ex}
\renewcommand{\arraystretch}{1.9}
\resizebox{0.48\textwidth}{!}{
\begin{tabular}{c|cccccc}
\toprule
\textbf{Model} & \ds{ Genius} & \ds{ ArXiv} & \ds{ Penn94} & \ds{ Twitch} & \ds{ Snap} & \ds{ Pokec} \\ 
\midrule
GCN-1 & 62.84 & 42.97 & 81.86 & 61.33 & 40.78 & 71.15 \\ 
GCN-2 & 61.84 & 42.93 & 81.90 & 61.28 & 41.34 & 72.78 \\
GCN-3 & 66.80 & 43.43 & 77.58 & 62.81 & 44.17 & 67.63 \\ 
SIGMA-1 & \textbf{91.41} & \textbf{55.41} & 85.27 & 67.29 & \textbf{64.71} & \textbf{82.24} \\ 
SIGMA-2 & 87.63 & 55.32 & \textbf{85.43} & \textbf{67.34} & 64.79 & 82.2 \\ 
SIGMA-3 & 86.67 & 54.91 & 85.27 & 67.09 & 64.69 & 82.19 \\ 
\bottomrule
\end{tabular}}
\end{table}
\subsection{Iterative Aggregation Mechanism Exploration}
In the previous sections, \aggname{} used a one-time aggregation mechanism with the SimRank matrix $\textbf{S}$, which, as shown in experiments, provides excellent performance. However, \aggname{} can also be adapted to iterative computations. In Eq.~(\ref{eq:zsim}), we perform a one-time MLP transformation and aggregation, but this design can be extended to other architectures and propagation schemes as a general edge rewiring method. For example, in an iterative GCN model, the aggregation steps are represented as: $\textbf{Z} = \sigma(...\sigma(\textbf{A}\ \sigma(\textbf{AXW})\textbf{W})...)$. By replacing $\textbf{A}$ with $\textbf{S}$, the aggregation process becomes: $\textbf{Z} = \sigma(...\sigma(\textbf{S}\ \sigma(\textbf{SXS})\ \textbf{W})...)$, where $\mathbf{\mathbf{}X_S} = \delta \mathbf{AW_A} + (1 - \delta)\mathbf{XW_X}$. We conducted exploratory experiments on SIGMA with iterative designs, as shown in Table~\ref{tab:iterative}. The results show that compared to GCN, SIGMA aggregation significantly improves performance, especially on heterophilous graphs, due to the combination of high-quality initial features $\mathbf{X_S}$ and the global aggregation from $S$. Omitting either component results in lower performance. Therefore, we consider \aggname{} a unique GNN model efficiently addressing heterophily graph problems.

{\color{black}
\section{Conclusion}
\label{sec:future}

In this paper, we propose \aggname{}, a novel heterophilous GNN model that leverages SimRank for efficient global aggregation, supported by rigorous theory and practical insights. In future, we will adapt \aggname{} to dynamic graphs by employing incremental and lazy update strategies for the SimRank matrix, which is cost for handling changes for \aggname{}, and to heterogeneous graphs by enhancing SimRank with type-aware adjacency and weighted aggregation to handle multiple node and edge types. We hope \aggname{} will invoke great insights for addressing heterophily graph learning problems and holds potential to inspire further research in other areas.
}

\section*{ACKNOWLEDGMENTS}
This research is supported by Singapore MOE AcRF Tier-2 Grant (T2EP20122-0003) and NTU-NAP startup grant (022029- 00001).

\clearpage
\newpage
\bibliographystyle{IEEEtranN}
\bibliography{reference}

\end{document}